\documentclass[11pt]{article}

\usepackage[final]{acl}
\usepackage{xspace}

\newcommand{\dataset}[1]{\textsc{#1}\xspace}

\newcommand{\feverous}{\dataset{FEVEROUS}}
\newcommand{\hybridqa}{\dataset{HybridQA}}
\newcommand{\sqa}{\dataset{SQA}}
\newcommand{\tabfact}{\dataset{TabFact}}

\newcommand{\ds}{\dataset{TabVerse}}

\newcommand{\html}{HTML}
\newcommand{\markdown}{Markdown}
\newcommand{\latex}{LaTeX}

\usepackage{times}
\usepackage{latexsym}
\usepackage{enumitem}

\usepackage[T1]{fontenc}

\usepackage[utf8]{inputenc}

\usepackage{microtype}
\usepackage{todonotes}
\usepackage{caption}
\usepackage{inconsolata}

\usepackage{graphicx}
\usepackage{booktabs}
\usepackage{float}
\usepackage{afterpage}
\usepackage{graphicx}
\usepackage{subcaption}
\usepackage{longtable,multirow}
\usepackage{tabularx}   
\usepackage{ragged2e}
\newcolumntype{Y}{>{\RaggedRight\arraybackslash}X} 
\usepackage{array}      
\renewcommand{\arraystretch}{1.08}
\usepackage{makecell}
\usepackage[table]{xcolor}
\usepackage{xcolor}
\usepackage{colortbl}
\usepackage{xparse}  
\usepackage{pgf}     
\usepackage{xcolor}
\usepackage{pifont}
\usepackage[most]{tcolorbox}
\usepackage{enumitem}
\usepackage{xcolor}
\usepackage{amsmath}
\usepackage{amssymb}
\usepackage{hyperref}
\usepackage{fontawesome5}

\usepackage{bbm}

\tcbset{
promptbox/.style={
    enhanced,
    breakable,
    colback=gray!8,
    colframe=black,
    boxrule=0.8pt,
    arc=2mm,
    left=6pt,
    right=6pt,
    top=6pt,
    bottom=6pt,
    title filled=false,
    coltitle=black,
    fonttitle=\bfseries,
    sharp corners=south,
    rounded corners=northwest,
    rounded corners=northeast
}}

\usepackage{siunitx}

\newcommand{\cmark}{\textcolor{green!60!black}{\ding{51}}} 
\newcommand{\xmark}{\textcolor{red!70!black}{\ding{55}}}   

\NewDocumentCommand{\colornum}{m}{%
  \ifdim#1pt<5.1pt%
    \colorbox{red!30}{#1}
  \else\ifdim#1pt<15.1pt%
    \colorbox{red!15}{#1}
  \else\ifdim#1pt<30.1pt%
    \colorbox{orange!25}{#1}
  \else\ifdim#1pt<50.1pt%
    \colorbox{yellow!25}{#1}
  \else\ifdim#1pt<70.1pt%
    \colorbox{lime!25}{#1}
  \else\ifdim#1pt<85.1pt%
    \colorbox{green!25}{#1}
  \else%
    \colorbox{green!40}{#1}
  \fi\fi\fi\fi\fi\fi%
}

\NewDocumentCommand{\colorqa}{m}{%
  \ifdim#1pt<30.1pt%
    \colorbox{red!20}{#1}
  \else\ifdim#1pt<50.1pt%
    \colorbox{orange!20}{#1}
  \else\ifdim#1pt<70.1pt%
    \colorbox{yellow!20}{#1}
  \else%
    \colorbox{green!20}{#1}
  \fi\fi\fi%
}

\NewDocumentCommand{\colorbleu}{m}{%
  \ifdim#1pt<2.1pt%
    \colorbox{red!20}{#1}
  \else\ifdim#1pt<4.1pt%
    \colorbox{orange!20}{#1}
  \else\ifdim#1pt<6.1pt%
    \colorbox{yellow!20}{#1}
  \else%
    \colorbox{green!20}{#1}
  \fi\fi\fi%
}

\NewDocumentCommand{\colormetric}{m}{%
  \ifdim#1pt<20.1pt%
    \colorbox{red!20}{#1}
  \else\ifdim#1pt<40.1pt%
    \colorbox{orange!20}{#1}
  \else\ifdim#1pt<60.1pt%
    \colorbox{yellow!20}{#1}
  \else\ifdim#1pt<80.1pt%
    \colorbox{lime!20}{#1}
  \else%
    \colorbox{green!20}{#1}
  \fi\fi\fi\fi%
}
\usepackage{cancel}
\usepackage{booktabs}
\usepackage{multirow}
\usepackage{colortbl}
\usepackage{xcolor}
\definecolor{headerpink}{RGB}{245,220,220}
\definecolor{headergreen}{RGB}{220,245,220}
\definecolor{headerblue}{RGB}{220,230,245}
\newcolumntype{C}[1]{>{\centering\arraybackslash}p{#1}}

\title{
    \raisebox{-0.2cm}{\includegraphics[width=0.8cm]{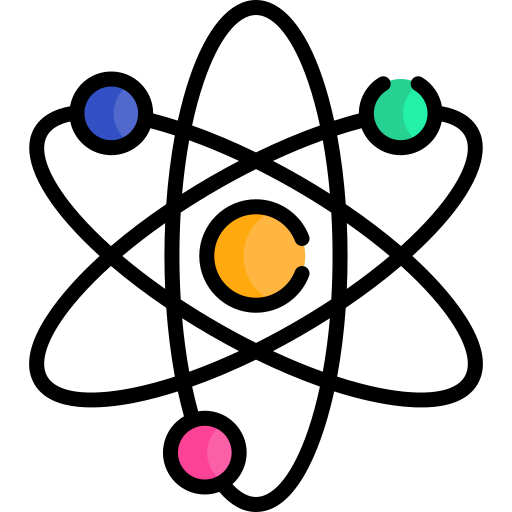}}  
    \hspace{0.01cm} \textbf{\Large \ds: Benchmarking Cross-Format Table\\Understanding in LLMs and VLMs}
}

\author{\textbf{Momina Ahsan}$^{1}$, \textbf{Sarfraz Ahmad}$^{1}$, \textbf{Ming Shan Hee}$^{1}$, \\ \textbf{Roy Ka-Wei Lee}$^{2}$, \textbf{Preslav Nakov}$^{1}$ \\
$^{1}$Mohamed bin Zayed University of Artificial Intelligence (MBZUAI) \\
$^{2}$Singapore University of Technology and Design (SUTD) \\
\parbox{\linewidth}{\centering
\texttt{\{momina.ahsan, preslav.nakov\}@mbzuai.ac.ae} \\
[0.3em]
\faGlobe\ \href{https://mbzuai-nlp.github.io/TABVERSE/}{Project}
\quad
\faDatabase\ \href{https://huggingface.co/datasets/MBZUAI/TABVERSE}{\ds{}}
\quad
\faGithub\ \href{https://github.com/mbzuai-nlp/TABVERSE}{Code}
\quad
\faTrophy\ \href{https://mbzuai-nlp.github.io/TABVERSE/leaderboard.html}{Leaderboard}
}
}

\begin{document}
\maketitle

\begin{abstract}
Large Language Models (LLMs) and Vision-Language Models (VLMs) are increasingly evaluated on table reasoning tasks, but the role of table representation remains under-explored. In practice, the same table content may appear in different structural formats, such as \html{}, \markdown{}, and \latex{}, or as rendered images. However, existing evaluations often let content, format, layout, and modality vary together, making it difficult to isolate representation effects. We introduce \ds{}, a controlled multimodal table benchmark that aligns the same table content across multiple structural formats and rendered images, with question category and difficulty tags. This design enables systematic evaluation of representation effects while holding table content fixed. We evaluate LLMs and VLMs across three tasks: Question Answering (QA), Structural Understanding Capability (SUC), and Structure Reconstruction (SR). Our results show that representation choice substantially affects table understanding. Models generally perform better with structured text than with rendered images, but the size of this gap depends on the task, model, and format. HTML is often the most robust text format, while row-sensitive structural tasks and syntactically usable LaTeX reconstruction remain challenging. These findings show that table representation is a key factor in reliable table evaluation.
\end{abstract}

\section{Introduction}
\label{sec:introduction}

Tables are widely used to present structured information in scientific documents, reports, and web content. This makes table understanding critical for AI systems that interpret and verify real-world data~\cite{smock2022pubtables}. Despite strong progress in Large Language Models (LLMs) and Vision-Language Models (VLMs), table comprehension remains challenging~\cite{brown2020language,touvron2023llama,bubeck2023sparks}.

Unlike plain text, tables require models to interpret both content and structure, including headers, merged cells, row and column boundaries, and relevant cells~\cite{deng-etal-2024-tables,sui2024table,kim2024tablevqa}.
The same table content can also be presented in different ways i.e., as \html{}, \latex{}, or \markdown{}, or as a rendered image in a PDF or screenshot. 
These representations expose different cues. Structured text provides markup and delimiters, while images provide visual layout, so model performance can change even when the underlying table content is identical.

Recent work has introduced table-specialized models~\cite{zhang-etal-2024-tablellama,deng2025rethinking} and shown that table reasoning is sensitive to serialization, prompting, and modality choices~\cite{deng-etal-2024-tables,sui2024table,sui2023tap4llm,singha2023tabular}. 
However, many benchmarks and evaluation pipelines let table content, format, layout, and modality vary together, making it difficult to isolate the effect of representation itself.

We introduce \ds{}, a benchmark for controlled cross-format and cross-modality table evaluation. 
\ds{} aligns identical tables across three structural formats (\html{}, \latex{}, \markdown{}) and their rendered images, enabling comparison while holding table content fixed. 
Built from held-out evaluation splits of \textsc{FEVEROUS}, \textsc{HybridQA}, \textsc{TabFact}, \textsc{SQA}, and \textsc{WikiTableQuestions}, it includes a full tagged pool and a 700-sample balanced evaluation set balanced by question category and difficulty.

We evaluate LLMs and VLMs on three complementary tasks: Question Answering (QA), Structural Understanding Capability (SUC), and Structure Reconstruction (SR). 
QA measures answer prediction under different table representations; SUC probes structure understanding through boundary detection, size estimation, and index-based retrieval; and SR measures whether VLMs can reconstruct tables from rendered images.

Our contributions are as follows:
\begin{itemize}[topsep=3pt,itemsep=3pt,parsep=0pt,partopsep=0pt]
    \item We formulate cross-format and cross-modality table understanding as a controlled evaluation problem, where table content is fixed while structural format and input modality vary under matched pipelines.

    \item We introduce \ds{}, an aligned multimodal table benchmark with \html{}, \latex{}, and \markdown{} representations, corresponding rendered images, category and difficulty tags, and a 700-sample balanced evaluation set from five TableQA sources.

    \item We benchmark LLMs and VLMs across matched text-only and image-based table inputs on QA, SUC, and SR, revealing how format and modality choices change model behavior across tasks and question groups, and how SR errors separate into table reconstruction quality and output usability.
\end{itemize}

Our experiments show that representation matters. Structured text often outperforms rendered images, HTML is often the most robust text format, and usable \latex{} reconstruction remains challenging.

\section{Related Work}
\label{sec:related-work}

\begin{table*}[t]
\centering
\small
\setlength{\tabcolsep}{1.2pt}
\renewcommand{\arraystretch}{0.72}
\definecolor{headerbg}{HTML}{FCE5CD}   
\definecolor{groupbg}{HTML}{D9EAF7}    
\definecolor{tierprop}{HTML}{FCE5CD}
\definecolor{tiertable}{HTML}{D9EAD3}

\begin{tabular}{@{}lcc cccc ccc@{}}
\toprule
&
\multicolumn{2}{c}{\cellcolor{groupbg}\textbf{Representations}} &
\multicolumn{4}{c}{\cellcolor{groupbg}\textbf{Visual Renders}} \\
\cmidrule(lr){2-3} \cmidrule(lr){4-7}

\rowcolor{headerbg}
\textbf{Literature} &
\textbf{Text} &
\textbf{Images} &
\textbf{\html{}} &
\textbf{\latex{}} &
\textbf{\markdown{}} &
\textbf{Others} &
\textbf{Q-diff} &
\textbf{Q-cat} &
\textbf{Aligned} \\
\midrule

\rowcolor{tiertable}
\multicolumn{10}{l}{\textbf{Table Reasoning and Multimodal Benchmarks}} \\

TableVQA-Bench~\cite{kim2024tablevqa}
&\cmark &\cmark &\cmark &\xmark &\xmark &\cmark &\xmark &\xmark &\cmark \\

MTabVQA~\cite{singh2025mtabvqa}
&\xmark &\cmark &\xmark &\xmark &\xmark &\cmark &\xmark &\cmark &\xmark \\

MMTabQA~\cite{mathur-etal-2024-knowledge}
&\cmark &\cmark &\xmark &\xmark &\xmark &\cmark &\xmark &\xmark &\xmark \\

MMTabQA~\cite{mathur-etal-2024-knowledge}
&\xmark &\cmark &\xmark &\xmark &\xmark &\cmark &\xmark &\cmark &\xmark \\

NeedleInATable~\cite{wang2026needleinatable}
&\cmark &\cmark &\xmark &\xmark &\xmark &\cmark &\xmark &\xmark &\cmark \\

TableVLM~\cite{chen-etal-2023-tablevlm}
&\cmark &\cmark &\cmark &\xmark &\xmark &\xmark &\xmark &\xmark &\cmark \\

\midrule

\rowcolor{tierprop}
\multicolumn{10}{l}{\textbf{Evaluation Frameworks / Controlled Studies}} \\

Tables as Texts or Images~\cite{deng-etal-2024-tables}
&\cmark &\cmark &\xmark &\xmark &\xmark &\cmark &\xmark &\xmark &\cmark \\

RealHiTBench~\cite{wu2025realhitbench}
&\cmark &\cmark &\xmark &\xmark &\xmark &\cmark &\xmark &\cmark &\cmark \\

LongTableBench~\cite{li-etal-2025-longtablebench}
&\cmark &\xmark &\xmark &\xmark &\xmark &\xmark &\xmark &\xmark &\xmark \\

Image2Struct~\cite{roberts2024image2struct}
&\xmark &\cmark &\cmark &\cmark &\xmark &\xmark &\xmark &\xmark &\xmark \\

\midrule

\textbf{\ds{} (Ours)}
&\cmark &\cmark &\cmark &\cmark &\cmark &- &\cmark &\cmark &\cmark \\

\bottomrule
\end{tabular}
\caption{Table understanding resources related to \ds{}. Under \emph{Visual Renders}, \html{}, \latex{}, and \markdown{} indicate whether the work provides or evaluates tables rendered from those source formats, while \emph{Others} covers other visual styles or image sources. \emph{Q-diff} and \emph{Q-cat} denote question difficulty and category annotations. \emph{Aligned} denotes paired textual tables and images with the same table content.}
\label{tab:table_reasoning_resources}
\end{table*}


\autoref{tab:table_reasoning_resources} summarizes related resources, which cover many table reasoning tasks but rarely isolate representation effects because content, format, layout, and modality often vary together.

\paragraph{Table reasoning benchmarks:} Early benchmarks established table QA over semi-structured tables, covering lookup, filtering, aggregation, and simple symbolic operations~\cite{pasupat-liang-2015-compositional,zhong2017wikisql}. 
Later datasets expanded to sequential QA, fact verification, multi-hop reasoning over tables and text, open-domain and multi-table QA, and table-grounded generation~\cite{sqa,tabfact,hybridqa,feverous,chen2021open,wu2025mmqa,totto,nan2022fetqa}. 
Other resources target numerical reasoning~\cite{chen2021finqa,zhu-etal-2021-tat}, hierarchical tables~\cite{cheng2022hitab}, long-context cell retrieval~\cite{wang2026needleinatable}, and complex or multilingual table understanding~\cite{zhu-etal-2025-tableeval}. 
These benchmarks provide important testbeds, but most evaluate a fixed representation or task setting; \ds{} instead tests the same table-question pairs across aligned textual formats and rendered images.

\paragraph{Representation and multimodal table evaluation:} Prior work shows that table reasoning depends strongly on serialization, prompting, segmentation, and modality. 
Table Meets LLM and TAP4LLM study prompting, sampling, augmentation, and structural decomposition~\cite{sui2024table,sui2023tap4llm}, while tables-as-text-versus-image comparisons show that representation choice can substantially change performance~\cite{deng-etal-2024-tables}. 
LongTableBench and RealHiTBench evaluate long or hierarchical tables under multiple input formats~\cite{li-etal-2025-longtablebench,wu2025realhitbench}; related studies examine source-sensitive table understanding, table-image modeling, and cross-domain evaluation behavior~\cite{yang2025doestablesourcematter,chen-etal-2023-tablevlm,borisova-etal-2025-table}; and multimodal benchmarks cover visual QA, semi-structured tables, rendered table images, and table-image retrieval~\cite{kim2024tablevqa,singh2025mtabvqa,mathur-etal-2024-knowledge,zheng-etal-2024-multimodal,titiya2025mmtbench,talmor2021multimodalqa,lompo2025visualtableqa,li-etal-2026-tarvir,xu-etal-2026-efficient}. These works motivate format-aware and image-based table evaluation, but often vary table source, layout, visual complexity, and representation together across different experimental settings; \ds{} holds table content fixed while systematically varying structural format and input modality.

\paragraph{Table reconstruction and table-focused modeling:} Beyond QA, table reconstruction and table-structure recognition studies extract structured representations from rendered tables or document images~\cite{roberts2024image2struct,li-etal-2020-tablebank}. 
This is related to our SR task, but prior work usually focuses on recognition accuracy or one target representation rather than reconstruction across aligned input and output formats.
Table-specialized pretraining and instruction tuning have been proposed for table manipulation, reasoning, and generation~\cite{herzig2020tapas,gong2020tablegpt,zhang-etal-2024-tablellama,zha2023tablegpt,li2024tablegpt,su2024tablegpt2largemultimodalmodel,zhang-etal-2025-tablellm,deng2025rethinking}; recent systems also combine OCR-style transcription with LLM reasoning for table VQA~\cite{guo2025talent} or add supervision through code-driven reasoning traces and structure-aware guidance~\cite{nguyen2026coretab,zhu2026disco}. These efforts are complementary to \ds{}, which evaluates LLMs and VLMs under controlled QA, SUC, and SR settings across aligned textual and visual table representations.
\section{\ds{}}
\label{sec:dataset}

\ds{} is constructed from filtered held-out splits of five TableQA datasets: \feverous~\cite{feverous}, \tabfact~\cite{tabfact}, \sqa~\cite{sqa}, \hybridqa~\cite{hybridqa}, and WikiTableQuestions (\textsc{wikitq})~\cite{pasupat-liang-2015-compositional}. 
We keep only single-table questions answerable from the table alone and check for overlap with the corresponding training splits where identifiers are available (Appendix~\ref{app:dataset-details}). 
This yields a full pool of 6{,}097 question--table pairs from 4{,}434 unique tables. 
After category and difficulty tagging, we select a 700-sample balanced evaluation set covering 629 unique tables (Table~\ref{tab:dataset_statistics}).

\subsection{Aligned Formats and Rendered Images}

For each table, we create \html{}, \markdown{}, and \latex{} representations and render one image from each. 
We adapt conversion utilities from \citet{sui2024table} and extend them for dataset artifacts such as missing values and special characters. Standardized font size, padding, and width keep textual and visual versions aligned by construction. Implementation details are in Appendix~\ref{app:dataset-details}.

\subsection{Question Category and Difficulty}

Each question is tagged with a \emph{question category} and a binary \emph{difficulty label}. The seven categories are Simple Lookup, Conditional Lookup, Multi-item Lookup, Aggregation/Arithmetic, Comparison/Extremum, and Binary Verification (single-step and multi-hop). Gemini-3-Flash-Preview assigns the initial category tags, which are manually reviewed and corrected where needed. 

Difficulty is estimated from zero-shot QA on rendered table images. GPT-5.2 and Gemini-3-Flash-Preview answer each question using the three aligned image renders produced from \html{}, \markdown{}, and \latex{}, giving six correctness indicators per question. Questions scoring 0--3 are labeled \emph{Hard}; those scoring 4--6 are labeled \emph{Easy}.

\subsection{Balanced Evaluation Set}

We balance the final evaluation set by difficulty and question category. 
It contains 700 question--table pairs: 350 Easy and 350 Hard questions, with 50 examples per category within each difficulty level. 
These questions reference 629 unique tables.

\subsection{Supported Tasks}
\label{subsection:supportedtask}

\ds{} supports three tasks: Question-Answering (QA), Structural Understanding Capability (SUC), and Structure Reconstruction (SR).

\paragraph{QA:} Given a question and a table as structured text or a rendered image, the model predicts an answer following the source dataset conventions.

\paragraph{SUC:} SUC probes table structure through boundary detection, size estimation, and index-based retrieval. 
We adapt the templates from \citet{sui2024table} and extend them with probes for table grounding and document-style tables. 
Prompt templates appear in Appendix~\ref{sec:prompts}.

\paragraph{SR:} Given a rendered table image, the model reconstructs the table in \html{}, \markdown{}, or \latex{}. 

\begin{table}[t]
\centering
\small
\setlength{\tabcolsep}{4pt}
\renewcommand{\arraystretch}{0.95}
\definecolor{headerbg}{HTML}{FCE5CD}
\begin{tabular}{lcc}
\toprule
\rowcolor{headerbg}
\textbf{Split} & \textbf{Question--table pairs} & \textbf{Unique tables} \\
\midrule
Full tagged pool & 6{,}097 & 4{,}434 \\
Balanced set & 700 & 629 \\
\bottomrule
\end{tabular}
\caption{Dataset statistics for \ds{}. The full tagged pool contains all filtered question-table pairs after category and difficulty tagging. The balanced evaluation set is used for all experiments, with equal coverage across difficulty and question category.}
\label{tab:dataset_statistics}
\end{table}
\label{sec:methodology}

\begin{figure*}[t]
\centering
\includegraphics[width=0.92\textwidth,
                 keepaspectratio,
                 trim=10mm 110mm 15mm 30mm,clip]{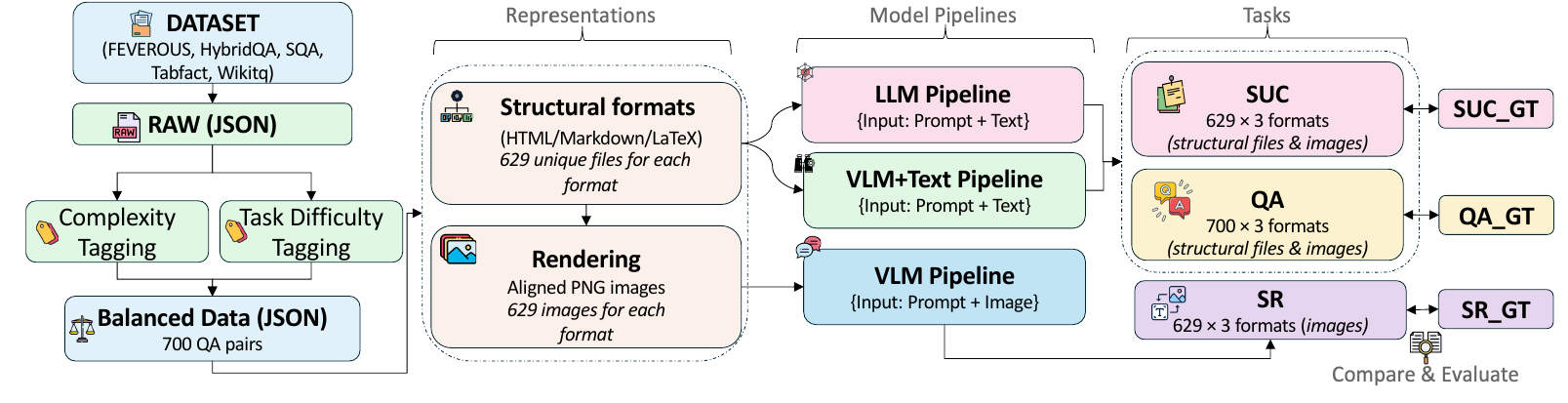}
\caption{Overview of \textbf{\ds}: From the balanced evaluation set, each table is represented in three structural formats (\html{}, \markdown{}, \latex{}) with corresponding rendered images. These aligned multimodal pairs enable evaluation on QA, SUC, and SR tasks across VLMs and LLMs for cross-format and cross-modality analysis.}
\label{fig:datasetconstruction}
\end{figure*}

\section{Experimental Settings}

We evaluate how structural format and input modality affect table understanding while keeping table content fixed. 
Using \ds{}, we vary the table format (\html{}, \markdown{}, \latex{}) and input modality (structured text vs.\ rendered image), allowing performance differences to be attributed to representation rather than content variation.

\subsection{Evaluation Pipelines}
\label{sec:pipelines}

We evaluate LLMs and VLMs on \ds{} using three different pipelines:

\textbf{VLM-Image:} The VLM receives each question with a \textit{rendered table image}. To measure visual format effects, we use three aligned image renderings per instance, rendered from \html{}, \markdown{}, and \latex{} sources, while keeping the question and table content unchanged.

\textbf{VLM-Text:} The VLM receives the question prompt with a \textit{structured text-based table} in one of the three formats, without an image. Comparing VLM-Text with VLM-Image isolates the impact of visual input within the same model.

\textbf{LLM-Text:} The LLM receives the question prompt with a \textit{structured text-based table} in one of the three formats. Comparing LLM-Text with VLM-Text highlights differences between language-only and multimodal models on identical text inputs.

\subsection{Models}

We evaluate several LLMs and VLMs, including general-purpose and table-specialized models. For LLMs, we use Qwen2.5-7B-Instruct~\cite{qwen25}, Qwen3-30B-A3B-Instruct~\cite{qwen3technicalreport}, TableGPT2-7B\footnotemark[1]~\cite{su2024tablegpt2largemultimodalmodel}, and TAMA-QWen3\footnotemark[1]~\cite{xing2025mmtu}.
For VLMs, we use SmolVLM2-2.2B-Instruct~\cite{marafioti2025smolvlm}, Gemma-3-12B-IT, Gemma-3-27B-IT~\cite{gemma3}, InternVL3.5-14B, InternVL3.5-30B-A3B~\cite{wang2025internvl3_5}, Qwen3-VL-8B-Instruct, Qwen3-VL-30B-A3B-Instruct~\cite{qwen3technicalreport}, Ministral-3-14B-Instruct~\cite{ministral3}, LLaVA-1.6-7B and LLaVA-1.6-13B~\cite{liu2023improved}, TableLLaVA-v1.5-7B\footnotemark[1]~\cite{zheng-etal-2024-multimodal}, GPT-5.2~\cite{gpt5}, and Gemini-3-Flash-Preview~\cite{gemini3flash}.
\footnotetext[1]{Table-specialized model for table understanding.}

\subsection{Evaluation Protocol}
\label{sec:models-configs}

All experiments follow a uniform zero-shot setup across the three tasks (Appendix~\ref{sec:prompts}).
Models generate outputs using greedy decoding (\texttt{temperature}=0, \texttt{top\_p}=1) with task-specific output limits: short for SUC, medium for QA, and long for SR. We apply minimal output normalization, including removing common answer prefixes such as \emph{the answer is} and normalizing whitespace and casing where appropriate. For QA and SUC, we report Exact-Match (EM) accuracy following the dataset conventions after light normalization. For SUC, we additionally report Field Accuracy and Relaxed Accuracy for pipe-separated structured answers. Field Accuracy compares each gold field with the prediction at the same position, while Relaxed Accuracy checks whether each gold field appears anywhere in the prediction. These are diagnostic metrics; exact match remains the primary SUC metric.


\begin{table*}[t]
\centering
\small
\setlength{\tabcolsep}{4pt}
\renewcommand{\arraystretch}{0.72}

\definecolor{headerbg}{HTML}{FCE5CD}
\definecolor{openbg}{HTML}{F4CCCC}
\definecolor{vlmbg}{HTML}{D9EAD3}
\definecolor{propbg}{HTML}{D9D2E9}
\definecolor{top}{HTML}{D9EAF7}

\begin{tabular}{l r ccc ccc}
\toprule
& &
\multicolumn{3}{c}{\cellcolor{top}\textbf{Table Image Render}} &
\multicolumn{3}{c}{\cellcolor{top}\textbf{Table Text Format}} \\
\cmidrule(lr){3-5}
\cmidrule(lr){6-8}
\rowcolor{headerbg}
\textbf{Model} & \textbf{Params.}
& \textbf{\html{}} & \textbf{LaTeX} & \textbf{\markdown{}}
& \textbf{\html{}} & \textbf{LaTeX} & \textbf{\markdown{}} \\
\midrule

\rowcolor{openbg}
\multicolumn{8}{l}{\textbf{Language Models (text-only)}}\\
Qwen2.5-IT~\cite{qwen25}                         & 7B        & --    & --    & --    & 44.57 & 42.71 & 45.43 \\
Qwen3-IT~\cite{qwen3technicalreport}             & 30B (A3B) & --    & --    & --    & \textbf{51.14} & \textbf{48.43} & \textbf{46.57} \\
TableGPT2~\cite{su2024tablegpt2largemultimodalmodel} & 7B    & --    & --    & --    & 44.43 & 41.57 & 42.14 \\
TAMA-QWen3~\cite{xing2025mmtu}                   & --        & --    & --    & --    & 18.29 & 19.14 & 20.71 \\
\addlinespace[4pt]

\rowcolor{vlmbg}
\multicolumn{8}{l}{\textbf{Vision-Language Models}}\\
SmolVLM2-IT~\cite{marafioti2025smolvlm}          & 2.2B      & 29.71 & 28.71 & 25.86 & 21.57 & 17.63 & 15.75 \\
Gemma-3-IT~\cite{gemma3}                         & 12B       & 38.86 & 39.57 & 38.57 & 50.29 & 49.00 & 48.57 \\
Gemma-3-IT~\cite{gemma3}                         & 27B       & 46.14 & 45.29 & 45.43 & \textbf{53.43} & 51.29 & 53.14 \\
InternVL3.5~\cite{wang2025internvl3_5}           & 14B       & 48.57 & 48.14 & 48.00 & 47.14 & 47.29 & 44.86 \\
InternVL3.5~\cite{wang2025internvl3_5}           & 30B (A3B) & 47.86 & \textbf{50.00} & 48.29 & 45.86 & 45.71 & 47.00 \\
Qwen3-VL-IT~\cite{qwen3vl}                       & 8B        & \textbf{50.29} & 49.29 & \textbf{49.71} & \textbf{53.43} & \textbf{52.14} & \textbf{53.29} \\
Qwen3-VL-IT~\cite{qwen3vl}                       & 30B (A3B) & 41.14 & 42.14 & 41.43 & 45.29 & 43.57 & 39.71 \\
Ministral-3-IT~\cite{ministral3}                 & 14B       & 44.43 & 39.14 & 42.71 & 40.00 & 35.43 & 36.57 \\
LLaVA-1.6$^*$~\cite{liu2023visualinstructiontuning}  & 7B        & 31.86 & 31.43 & 32.00 & 27.37 & 29.55 & 26.50 \\
LLaVA-1.6$^*$~\cite{liu2023visualinstructiontuning}  & 13B       & 25.14 & 23.71 & 25.00 & 23.91 & 22.13 & 24.89 \\
TableLLaVA-v1.5$^*$~\cite{zheng-etal-2024-multimodal} & 7B       &  1.29 &  1.00 &  4.00 & 23.61 & 27.37 & 28.40 \\

\addlinespace[4pt]
\rowcolor{propbg}
\multicolumn{8}{l}{\textbf{Proprietary Models}}\\
GPT-5.2~\cite{gpt5}                              & --        & 54.57 & 54.52 & 56.14 & 57.43 & 57.29 & 58.00 \\
Gemini-3-Flash-Preview~\cite{gemini3flash}       & --        & \textbf{65.43} & \textbf{65.14} & \textbf{65.43}
                                                                  & \textbf{65.71} & \textbf{65.00} & \textbf{65.43} \\
\bottomrule
\end{tabular}

\caption{
\textbf{TaskQA results (700 questions).}
EM accuracy (\%) across three aligned table representations (\html{}/LaTeX/\markdown{}).
Text-only models (LLM-Text) operate on structured table text (right block). Vision-language models are evaluated both on rendered table images (left block) and on structured table text (right block), which isolates modality effects while keeping the underlying table content identical.}
\label{tab:taskqa}
\end{table*}

For SR, we report \textsc{GriTS}~\cite{grits} using both GriTS-Topology and GriTS-Content.
GriTS-Topology measures structural similarity between the reconstructed and reference tables, while GriTS-Content measures cell-text fidelity.
We also measure syntactic usability for each requested target format using HTML parse success, Markdown render/parse success, and LaTeX compilation success. To connect usability with reconstruction quality, Appendix~\ref{app:sr} reports two usability-aware variants of \textsc{GriTS}. 
Valid-only \textsc{GriTS} averages scores only over syntactically usable outputs when at least one output is usable:

\begin{equation}
\mathrm{GriTS}_{\mathrm{valid}} =
\frac{\sum_{i=1}^{N} u_i \cdot \mathrm{GriTS}_i}
{\sum_{i=1}^{N} u_i}
\label{eq:grits_valid}
\end{equation}

where $u_i=1$ if the generated output is syntactically usable and $u_i=0$ otherwise.
Zero-penalized \textsc{GriTS} instead assigns unusable outputs a score of zero before averaging:

\begin{equation}
\mathrm{GriTS}_{0} =
\frac{1}{N}\sum_{i=1}^{N} u_i \cdot \mathrm{GriTS}_i
\label{eq:grits_zero}
\end{equation}

\begin{table*}[t]
\centering
\small
\setlength{\tabcolsep}{3.2pt}
\renewcommand{\arraystretch}{0.72}

\definecolor{headerbg}{HTML}{FCE5CD}
\definecolor{openbg}{HTML}{F4CCCC}
\definecolor{propbg}{HTML}{D9D2E9}

\begin{tabular}{l c c c c c c c c c c c c}
\toprule
\rowcolor{headerbg}
\textbf{Models} & \textbf{Formats} &
\textbf{T.P.} & \textbf{F.C.} & \textbf{L.C.} &
\textbf{S.D.} & \textbf{\# Rows} & \textbf{\# Cols} &
\textbf{C.Lu.} & \textbf{R.Lu.} & \textbf{Co.Rt.} & \textbf{Ro.Rt.} &
\textbf{Overall} \\
\midrule

\rowcolor{openbg}
\multicolumn{13}{l}{\textbf{Open Models}} \\

\multirow{3}{*}{Gemma-3-27B-IT}
 & \html{}     & 29.6 & 63.9 & 52.3 & 21.0 & 39.9 & 50.1 & 22.6 & 32.6 & 70.9 & 10.5 & 39.3 \\
 & LaTeX       & 28.3 & 59.6 & 52.0 & 23.5 & 39.7 & 50.2 & 22.7 & 32.6 & 70.3 & 9.1  & 38.8 \\
 & \markdown{} & 30.2 & 59.0 & 51.5 & 27.3 & 44.4 & 55.8 & 24.2 & 33.7 & 74.2 & \textbf{11.3} & 41.2 \\

\multirow{3}{*}{InternVL3.5-14B}
 & \html{}     & 35.3 & 86.8 & 68.4 & 22.3 & 55.3 & 89.8 & 15.6 & 15.9 & 84.6 & 4.1 & 47.8 \\
 & LaTeX       & 38.0 & 88.6 & 71.9 & 16.4 & 28.6 & 83.1 & 18.9 & 22.7 & 90.1 & 6.4 & 46.5 \\
 & \markdown{} & 31.6 & 83.0 & 65.3 & 32.3 & 51.0 & 86.6 & 20.3 & 23.5 & 88.9 & 5.4 & 48.8 \\

\multirow{3}{*}{Qwen3-VL-30B-A3B-IT}
 & \html{}     & 50.2 & 92.5 & 85.2 & 32.9 & 44.2 & 91.9 & 21.1 & 24.3 & 71.1 & 0.2 & 51.4 \\
 & LaTeX       & 45.6 & 92.7 & 82.2 & 34.2 & 40.2 & 87.4 & 22.3 & 27.3 & 83.8 & 4.0 & 52.0 \\
 & \markdown{} & 40.1 & 87.9 & 78.7 & 32.0 & 41.7 & 87.4 & 21.6 & 25.1 & 88.9 & 1.1 & 50.4 \\

\multirow{3}{*}{Ministral-3-14B-IT}
 & \html{}     & 36.1 & 40.5 & 61.5 & 22.7 & 35.1 & 46.4 & 26.6 & \textbf{39.6} & 79.7 & 10.7 & 39.9 \\
 & LaTeX       & 36.1 & 41.8 & 53.3 & 38.2 & 35.6 & 36.1 & \textbf{26.9} & 39.4 & 72.7 & 11.1 & 39.1 \\
 & \markdown{} & 30.7 & 34.3 & 49.4 & 38.5 & 35.6 & 52.5 & 26.6 & 38.8 & 74.2 & 10.7 & 39.1 \\

\multirow{3}{*}{LLaVA-1.6-13B}
 & \html{}     & 0.2 & 31.5 & 27.5 & 4.5 & 19.4 & 28.0 & 1.4 & 6.4 & 37.5 & 0.0 & 15.6 \\
 & LaTeX       & 0.2 & 32.9 & 29.6 & 3.8 & 20.0 & 20.5 & 2.2 & 5.6 & 36.7 & 0.2 & 15.2 \\
 & \markdown{} & 0.5 & 35.6 & 31.6 & 4.3 & 19.1 & 22.6 & 2.1 & 6.2 & 36.6 & 0.0 & 15.9 \\

\addlinespace[4pt]
\rowcolor{propbg}
\multicolumn{13}{l}{\textbf{Proprietary Models}} \\

\multirow{3}{*}{GPT-5.2}
 & \html{}     & \textbf{93.0} & \textbf{97.1} & \textbf{94.9} & 32.9 & 78.4 & 98.1 & 2.7 & 15.7 & 95.2 & 1.7 & 61.0 \\
 & LaTeX       & 85.4 & 93.3 & 88.1 & 57.7 & 85.7 & 96.5 & 3.5 & 19.2 & 93.5 & 2.5 & 62.5 \\
 & \markdown{} & 87.4 & 93.6 & 91.6 & \textbf{80.3} & \textbf{91.3} & 98.9 & 3.0 & 20.8 & 96.0 & 5.1 & \textbf{66.8} \\

\multirow{3}{*}{Gemini-3-Flash-Preview}
 & \html{}     & 91.7 & 97.0 & 88.7 & 0.2 & 0.5 & 94.9 & 0.6 & 14.8 & 94.4 & 0.3 & 48.3 \\
 & LaTeX       & 86.1 & 94.8 & 87.4 & 0.0 & 6.5 & 99.2 & 1.6 & 15.1 & 95.7 & 0.8 & 48.7 \\
 & \markdown{} & 87.9 & 94.8 & 87.7 & 0.2 & 1.9 & \textbf{99.7} & 1.3 & 14.9 & \textbf{97.3} & 0.3 & 48.6 \\

\bottomrule
\end{tabular}

\caption{
\textbf{SUC results for the VLM-Image pipeline.}
EM accuracy (\%) across ten structure-oriented subtasks, including table partitioning (T.P.), boundary detection (F.C., L.C.), size estimation (S.D., \#Rows, \#Cols), coordinate lookup (C.Lu., R.Lu.), and index-based retrieval (Co.Rt., Ro.Rt.).
Models receive rendered table images derived from \html{}, LaTeX, and \markdown{} sources.
}
\label{tab:suc_vlm_image}
\end{table*}

\begin{table}[t]
\centering
\small
\setlength{\tabcolsep}{4pt}
\renewcommand{\arraystretch}{0.85}

\definecolor{headerbg}{HTML}{FCE5CD}
\definecolor{deltabg}{HTML}{D9EAF7}

\begin{tabular}{lcccc}
\toprule
&
\multicolumn{4}{c}{\cellcolor{deltabg}\textbf{$\Delta$ (Field Acc. - EM)}} \\
\cmidrule(lr){2-5}
\rowcolor{headerbg}
\textbf{Model} &
 \textbf{T.P.} & \textbf{S.D.} & \textbf{C.Lu.} & \textbf{Ro.Rt.} \\
\midrule
Gemma-3-27B-IT              & +25.7 & +21.3 & +20.9 & +13.7 \\
InternVL3.5-14B             & +30.8 & +29.5 & +30.2 & +11.8 \\
Qwen3-VL-30B-A3B-IT         & +28.3 & +30.1 & +27.0 & +20.6 \\
Ministral-3-14B-IT          & +18.2 & +25.5 & +20.2 & +11.7 \\
LLaVA-1.6-13B               &  +6.2 & +17.2 &  +7.8 &  +3.6 \\
GPT-5.2                     &  +4.9 & +21.0 & +42.1 & +13.1 \\
Gemini-3-Flash-Preview      &  +7.3 & +48.9 & +35.2 & +13.4 \\
\bottomrule
\end{tabular}

\caption{
\textbf{Field-level gaps on selected SUC subtasks.}
Values report $\Delta = \mathrm{Field\ Accuracy} - \mathrm{EM}$, averaged over \html{}, LaTeX, and \markdown{} image renders.
Larger gaps indicate subtasks where models often recover part of the structured answer but fail exact match.
}
\label{tab:suc_field_gap}
\end{table}

\begin{table}[t]
\centering
\small
\setlength{\tabcolsep}{1pt}
\renewcommand{\arraystretch}{0.85}

\definecolor{headerbg}{HTML}{FCE5CD}
\definecolor{deltabg}{HTML}{D9EAF7}

\begin{tabular}{lrrrrrr}
\toprule
&
\multicolumn{6}{c}{\cellcolor{deltabg}\textbf{$\Delta_{\mathrm{EM}}$ (Explicit -- Implicit)}} \\
\cmidrule(lr){2-7}
\rowcolor{headerbg}
\textbf{Model} &
\textbf{T.P.} &
\textbf{F.C.} &
\textbf{C.Lu.} &
\textbf{R.Lu.} &
\textbf{Co.Rt.} &
\textbf{Ro.Rt.} \\
\midrule
Gemma-3-27B-IT              & +13.2 & +28.4 & +22.0 &  +7.0 &  +9.7 &  +4.1 \\
InternVL3.5-14B             & +17.0 & +65.8 & +18.0 & +15.9 & +26.7 &  +1.7 \\
Qwen3-VL-30B$^*$     & +33.1 & +79.6 & +21.7 &  +9.8 &  +5.1 &  +0.0 \\
Ministral-3-14B-IT          &  -8.2 &  +6.6 & +24.4 & +23.6 &  +0.3 &  +3.6 \\
LLaVA-1.6-13B          &  +0.2 & -21.1 &  +1.6 &  -0.7 &  +3.2 &  +0.1 \\
GPT-5.2                     & +80.5 & +83.8 &  +1.4 & -52.7 & +16.8 &  -1.9 \\
Gemini$^*$     & +88.3 & +94.8 &  +0.7 & -15.4 &  -0.7 &  +0.5 \\
\bottomrule
\end{tabular}

\caption{
\textbf{Effect of prompt explicitness on SUC performance.} Values report $\Delta_{\mathrm{EM}}=\mathrm{EM}_{\mathrm{explicit}}-\mathrm{EM}_{\mathrm{implicit}}$ for VLMs, averaged across \html{}, \latex{}, and \markdown{} renders. Positive values favor explicit prompts, while negative values favor implicit prompts. Qwen3-VL-30B$^*$ denotes Qwen3-VL-30B-A3B-IT, and Gemini$^*$ denotes Gemini-3-Flash-Preview.}
\label{tab:prompt_sensitivity_changed_vlm}
\end{table}

\section{Results}
\label{sec:results}
This section evaluates model performance across different table formats, input modalities, and table comprehension tasks.

\subsection{Question Answering}
\label{sec:taskqa}

Table~\ref{tab:taskqa} reports EM accuracy on 700 QA questions across three aligned table representations.

\paragraph{Overall performance:} Gemini-3-Flash-Preview obtains the highest scores across all formats and modalities, followed by GPT-5.2.
Among open-weight VLMs, Qwen3-VL-8B-IT is strongest under strict EM, while Qwen3-30B-A3B-IT is the strongest text-only LLM. Larger models are not always better under strict EM: the 8B Qwen3-VL variant outperforms the 30B-A3B variant across modalities and formats in Table~\ref{tab:taskqa}.
However, Appendix~\ref{tab:taskqa_em_relaxed_all} shows that the 30B-A3B variant recovers substantially under relaxed matching, especially in VLM-Text. This suggests that part of the strict EM gap comes from answer-formatting behavior rather than answer retrieval alone. InternVL3.5 shows a milder version of the same pattern.

\paragraph{Modality and format effects:} Structured table text does not consistently outperform rendered images.
Gemma-3 benefits from text inputs, SmolVLM2 and InternVL3.5 perform similarly or better on images, and Gemini-3-Flash-Preview remains nearly unchanged across modalities. At the format level, rendered-image scores are generally similar across \html{}, \latex{}, and \markdown{}, while text pipelines show larger gaps, suggesting that symbolic table format affects text-based reasoning more than rendered-image reasoning.

\paragraph{Task factors:} Beyond modality and format, QA performance varies by question category and difficulty.
Verification questions tend to be easier, while multi-item lookup and aggregation/counting questions remain challenging, as illustrated by the category-wise profiles in Figure~\ref{fig:task-category-radar}.
Accuracy also drops substantially from Easy to Hard questions. We provide the full category and difficulty breakdowns in Appendix~\ref{app:taskqa}.


\subsection{Structural Understanding Capability}
\label{sec:suc}

Tables~\ref{tab:suc_vlm_image}--\ref{tab:prompt_sensitivity_changed_vlm} evaluate SUC across ten structure-oriented subtasks defined in Section~\ref{subsection:supportedtask}. Tables~\ref{tab:suc_vlm_image} and~\ref{tab:suc_vlm_image_appendix} report VLM-Image EM results, while Tables~\ref{tab:suc_vlm_text} and~\ref{tab:suc_llm_text} report VLM-Text and LLM-Text results.

\paragraph{Overall performance:} GPT-5.2 obtains the highest overall VLM-Image scores, while Qwen3-VL-30B and InternVL3.5-14B are the strongest open-weight VLMs in Table~\ref{tab:suc_vlm_image}. Gemini-3-Flash performs well on boundary and column-oriented subtasks but struggles with size detection, cell lookup, and row retrieval, making SUC performance highly task-dependent. Column counting and column retrieval are among the easiest subtasks, whereas row retrieval, cell lookup, table partitioning, and size detection remain difficult, as shown in Figure~\ref{fig:suc_task_difficulty}. This gap is most visible in row-oriented reasoning, where even strong models fail to retrieve the correct indexed row.

Format sensitivity is modest in the VLM-Image setting. For most models, overall scores vary by only a few points across \html{}, \latex{}, and \markdown{} renders, suggesting that visual table structure largely dominates source-format differences after rendering. GPT-5.2 is the main exception, improving from 61.0 on \html{} to 66.8 on \markdown{}, driven mainly by stronger size detection and row-count estimation. Boundary-related subtasks (F.C., L.C.) are generally solved reliably, with GPT-5.2, Gemini-3-Flash, and Qwen3-VL-30B exceeding 85\% on most formats, whereas coordinate-based reasoning remains much harder. Row retrieval stays below 12\% for all open models and below 6\% even for the strongest proprietary models, highlighting a persistent gap between recognizing table structure and accurately navigating row-level coordinates.

\paragraph{Field Accuracy reveals partial structural recovery:} Field Accuracy shows that low EM often reflects incomplete or shifted localization rather than entirely wrong table understanding. In Table~\ref{tab:suc_field_gap}, the largest gains occur on row/column-sensitive subtasks: InternVL3.5-14B gains +30.8 on table partitioning and +30.2 on cell lookup, while Qwen3-VL-30B gains +30.1 on size detection. These gaps are especially informative for row-sensitive outputs, where models often recover part of the structure but miss exact localization.

\paragraph{Header and indexing cues affect SUC scores:} Table~\ref{tab:prompt_sensitivity_changed_vlm} compares the explicit SUC prompt, which states header exclusion and 0-indexed row/column coordinates, with an implicit prompt that removes these details. The effect is strongest on index-dependent subtasks. For first-cell detection, the explicit prompt gives much higher EM for GPT-5.2 (+83.8) and Gemini-3-Flash (+94.8), with similar gains for the strongest open models; table partitioning shows the same trend for GPT-5.2 (+80.5) and Gemini-3-Flash (+88.3). The effect is much weaker for last-cell detection, suggesting that many first-cell errors come from treating the header as the first row. In contrast, reverse lookup improves under the implicit prompt for GPT-5.2 (-52.7) and Gemini-3-Flash (-15.4). Thus, SUC also tests whether models follow the intended row-indexing and header-inclusion convention.

\paragraph{Pipeline and format effects:} Tables~\ref{tab:suc_vlm_image}, \ref{tab:suc_vlm_image_appendix}, \ref{tab:suc_vlm_text}, and~\ref{tab:suc_llm_text} show that SUC is generally stronger with structured table text than with rendered images. 

The gains are clearest for GPT-5.2 and Gemini-3-Flash, while open-weight models improve less uniformly. The largest image-to-text improvements occur on row-boundary and header-sensitive tasks. Row retrieval and cell lookup remain the primary bottlenecks across pipelines, despite strong column counting and retrieval, though several VLMs improve on these tasks in the VLM-Text setting (Figure~\ref{fig:suc_pipeline_comparison}). Table~\ref{tab:suc_llm_text} shows that text alone does not solve SUC for LLMs. Text-only Qwen3-30B performs strongly on \html{} but drops on \latex{} and \markdown{}, while smaller and table-specialized LLMs remain weak on coordinate lookup and row retrieval, indicating format effects that are present but not universal. \html{} is often the safest text format for text-input pipelines, while rendered-image results show smaller and less consistent format differences. Overall, SUC depends on input modality, table format, and row/column indexing behavior.

\subsection{Structure Reconstruction}
\label{sec:sr}

Table~\ref{tab:grits_image_results} evaluates SR, where models reconstruct rendered table images into a given text format. We report \textsc{GriTS}-Topology, \textsc{GriTS}-Content, and output usability (Tables~\ref{tab:sr_output_usability} and~\ref{tab:sr_output_usability_app}). Appendix~\ref{app:sr} also reports valid-only and zero-penalized \textsc{GriTS} to distinguish usability failures from reconstruction errors.

\paragraph{SR errors reflect both structure and content:} Across models and formats, \textsc{GriTS}-Topology consistently exceeds \textsc{GriTS}-Content, showing that models recover table layout more reliably than exact cell text. Strong VLMs such as Qwen3, InternVL3.5, and GPT-5.2 obtain high topology scores across most source and target formats, but content scores drop more often, especially for \latex{} targets. This indicates that SR failures stem from both structural errors and cell-text degradation during reconstruction.

\paragraph{Usability exposes syntax-level failures:} Tables~\ref{tab:sr_output_usability} and~\ref{tab:sr_output_usability_app} show that high \textsc{GriTS} does not always imply usable output. Strong open VLMs usually produce valid \html{} and \markdown{}, but \latex{} usability is less stable. For example, Qwen3-VL-30B reaches perfect usability for \html{} and \markdown{} targets, while its \latex{} usability ranges from 0.77 to 0.95. This separates unusable syntax from usable but inaccurate reconstructions. We therefore report zero-penalized \textsc{GriTS} in Table~\ref{tab:sr_zero_penalized_grits}, where unusable outputs receive zero before averaging.

\begin{table*}[t]
\centering
\small
\setlength{\tabcolsep}{0.9pt}
\renewcommand{\arraystretch}{1}

\definecolor{tierprop}{HTML}{FCE5CD}   
\definecolor{tieropen}{HTML}{F4CCCC}   
\definecolor{tiertable}{HTML}{D9EAD3}  
\definecolor{deltabg}{HTML}{D9EAF7}    

\begin{tabular}{l *{18}{c}}
\toprule

& \multicolumn{6}{c}{\cellcolor{deltabg}\textbf{\html{} image}} &
\multicolumn{6}{c}{\cellcolor{deltabg}\textbf{\markdown{} image}} &
\multicolumn{6}{c}{\cellcolor{deltabg}\textbf{\latex{} image}} \\

\cmidrule(lr){2-7}\cmidrule(lr){8-13}\cmidrule(lr){14-19}

& \multicolumn{3}{c}{\cellcolor{deltabg}\textbf{\texttt{GriTS}-Topology}} & \multicolumn{3}{c}{\cellcolor{deltabg}\textbf{\texttt{GriTS}-Content}}
& \multicolumn{3}{c}{\cellcolor{deltabg}\textbf{\texttt{GriTS}-Topology}} & \multicolumn{3}{c}{\cellcolor{deltabg}\textbf{\texttt{GriTS}-Content}}
& \multicolumn{3}{c}{\cellcolor{deltabg}\textbf{\texttt{GriTS}-Topology}} & \multicolumn{3}{c}{\cellcolor{deltabg}\textbf{\texttt{GriTS}-Content}} \\

\cmidrule(lr){2-4}\cmidrule(lr){5-7}
\cmidrule(lr){8-10}\cmidrule(lr){11-13}
\cmidrule(lr){14-16}\cmidrule(lr){17-19}

\rowcolor{tierprop}
\textbf{Models}
& \textbf{\html{}} & \textbf{Md} & \textbf{TeX} & \textbf{\html{}} & \textbf{Md} & \textbf{TeX}
& \textbf{\html{}} & \textbf{Md} & \textbf{TeX} & \textbf{\html{}} & \textbf{Md} & \textbf{TeX}
& \textbf{\html{}} & \textbf{Md} & \textbf{TeX} & \textbf{\html{}} & \textbf{Md} & \textbf{TeX} \\
\midrule

\rowcolor{tieropen}
\multicolumn{19}{l}{\textbf{Open Models}} \\

SmolVLM2-2.2B & 0.87 & 0.86 & 0.81 & 0.74 & 0.74 & 0.69 & 0.66 & 0.79 & 0.78 & 0.57 & 0.68 & 0.68 & 0.79 & 0.85 & 0.57 & 0.65 & 0.69 & 0.47 \\
Gemma3-12B & 0.94 & 0.95 & 0.91 & 0.79 & 0.79 & 0.76 & 0.95 & 0.96 & 0.92 & 0.80 & 0.81 & 0.77 & 0.94 & 0.95 & 0.90 & 0.78 & 0.77 & 0.75 \\
Gemma3-27B & 0.97 & 0.97 & 0.94 & 0.86 & 0.85 & 0.82 & 0.98 & 0.97 & 0.94 & 0.85 & 0.85 & 0.81 & 0.97 & 0.96 & 0.94 & 0.83 & 0.82 & 0.81 \\
InternVL3.5-14B & \textbf{0.99} & 0.99 & \textbf{0.96} & 0.95 & 0.94 & \textbf{0.92} & \textbf{0.99} & 0.98 & 0.96 & 0.93 & 0.93 & 0.91 & 0.96 & 0.97 & 0.95 & 0.93 & 0.91 & 0.93 \\
InternVL3.5-30B & 0.98 & 0.99 & 0.95 & 0.95 & 0.94 & \textbf{0.92} & \textbf{0.99} & 0.99 & 0.96 & 0.93 & 0.94 & 0.92 & 0.96 & 0.97 & 0.95 & 0.93 & 0.92 & 0.92 \\
Qwen3-VL-8B & \textbf{0.99} & \textbf{1.00} & 0.95 & \textbf{0.98} & \textbf{0.98} & \textbf{0.92} & \textbf{0.99} & \textbf{1.00} & \textbf{0.97} & \textbf{0.97} & \textbf{0.98} & \textbf{0.93} & \textbf{0.98} & \textbf{0.98} & \textbf{0.97} & \textbf{0.95} & \textbf{0.95} & \textbf{0.95} \\
Qwen3-VL-30B & 0.98 & 0.99 & 0.84 & \textbf{0.98} & \textbf{0.98} & 0.80 & \textbf{0.99} & \textbf{1.00} & \textbf{0.97} & \textbf{0.97} & \textbf{0.98} & \textbf{0.93} & 0.97 & \textbf{0.98} & 0.96 & \textbf{0.95} & \textbf{0.95} & \textbf{0.95} \\
Ministral3-14B & 0.98 & 0.95 & 0.95 & 0.94 & 0.90 & \textbf{0.92} & 0.98 & 0.95 & 0.95 & 0.92 & 0.90 & 0.90 & 0.95 & 0.93 & 0.93 & 0.87 & 0.86 & 0.88 \\
LLaVA1.6-7B & 0.70 & 0.66 & 0.05 & 0.43 & 0.40 & 0.02 & 0.70 & 0.54 & 0.04 & 0.46 & 0.35 & 0.02 & 0.71 & 0.58 & 0.05 & 0.44 & 0.37 & 0.03 \\
LLaVA1.6-13B & 0.66 & 0.81 & 0.23 & 0.46 & 0.51 & 0.15 & 0.62 & 0.75 & 0.22 & 0.45 & 0.51 & 0.15 & 0.65 & 0.76 & 0.27 & 0.46 & 0.49 & 0.19 \\

\addlinespace[4pt]
\rowcolor{tierprop}
\multicolumn{19}{l}{\textbf{Proprietary Models}} \\

GPT-5.2 & 0.98 & 0.98 & 0.81 & 0.97 & 0.94 & 0.78 & 0.98 & 0.99 & 0.95 & 0.96 & 0.97 & 0.91 & \textbf{0.98} & 0.97 & 0.93 & 0.94 & 0.94 & 0.89 \\
Gemini-3-Flash & 0.05 & 0.96 & 0.65 & 0.05 & 0.94 & 0.64 & 0.86 & 0.97 & 0.51 & 0.85 & 0.96 & 0.51 & 0.65 & 0.93 & 0.58 & 0.63 & 0.91 & 0.57 \\

\addlinespace[4pt]
\rowcolor{tiertable}
\multicolumn{19}{l}{\textbf{Table-specialised Models}} \\

TableLLaVA-7B & 0.73 & 0.71 & 0.68 & 0.33 & 0.33 & 0.31 & 0.73 & 0.72 & 0.70 & 0.43 & 0.43 & 0.41 & 0.58 & 0.57 & 0.54 & 0.29 & 0.29 & 0.28 \\

\bottomrule
\end{tabular}

\caption{
\textbf{SR from table images.}
Models reconstruct table images rendered from \html{}, \markdown{}, or \latex{} into \html{}, \markdown{}, or \latex{}. We report \texttt{GriTS}-Topology (structure) and \texttt{GriTS}-Content (cell text); higher scores indicate better reconstruction. Best scores per column are shown in bold.}
\label{tab:grits_image_results}
\end{table*}

\begin{table}[t]
\centering
\scriptsize
\setlength{\tabcolsep}{0.6pt}
\renewcommand{\arraystretch}{1}

\definecolor{tieropen}{HTML}{F4CCCC}     
\definecolor{tierprop}{HTML}{FCE5CD}     
\definecolor{tiertable}{HTML}{D9EAD3}    
\definecolor{deltabg}{HTML}{D9EAF7}      

\begin{tabular}{l *{9}{c}}
\toprule

& \multicolumn{3}{c}{\cellcolor{deltabg}\textbf{\html{} image}} &
\multicolumn{3}{c}{\cellcolor{deltabg}\textbf{\markdown{} image}} &
\multicolumn{3}{c}{\cellcolor{deltabg}\textbf{\latex{} image}} \\

\cmidrule(lr){2-4}\cmidrule(lr){5-7}\cmidrule(lr){8-10}

\rowcolor{tierprop}
\textbf{Models}
& \textbf{\html{}} & \textbf{Md} & \textbf{TeX}
& \textbf{\html{}} & \textbf{Md} & \textbf{TeX}
& \textbf{\html{}} & \textbf{Md} & \textbf{TeX} \\
\midrule

\rowcolor{tieropen}
\multicolumn{10}{l}{\textbf{Open Models}} \\

Gemma3-27B-IT & \textbf{1.00} & \textbf{1.00} & 0.89 & \textbf{1.00} & \textbf{1.00} & 0.88 & \textbf{1.00} & \textbf{1.00} & 0.87 \\
Qwen3-VL-30B-A3B-IT & \textbf{1.00} & \textbf{1.00} & 0.77 & \textbf{1.00} & \textbf{1.00} & 0.89 & \textbf{1.00} & \textbf{1.00} & 0.95 \\
LLaVA1.6-Vicuna-13B & 0.99 & 0.90 & 0.00 & 0.93 & 0.84 & 0.00 & 0.99 & 0.90 & 0.01 \\
TableLLaVA-v1.5-7B & \textbf{1.00} & 0.96 & 0.71 & \textbf{1.00} & 0.97 & 0.74 & \textbf{1.00} & 0.98 & 0.80 \\

\addlinespace[4pt]
\rowcolor{tierprop}
\multicolumn{10}{l}{\textbf{Proprietary Models}} \\

GPT-5.2 & 0.99 & \textbf{1.00} & \textbf{0.94} & 0.99 & \textbf{1.00} & \textbf{0.96} & \textbf{1.00} & \textbf{1.00} & \textbf{0.98} \\
Gemini-3-Flash-Preview & 0.06 & 0.99 & 0.76 & 0.91 & \textbf{1.00} & 0.80 & 0.72 & 0.99 & 0.75 \\

\bottomrule
\end{tabular}

\caption{
\textbf{Output usability for SR.} Values report the fraction of syntactically usable reconstructed outputs across source render formats and target output formats.
Best scores per column are shown in bold.
}
\label{tab:sr_output_usability}
\end{table}

\paragraph{\latex{} and content preservation remain the main bottlenecks:} \latex{} reconstruction is difficult in two ways: models must recover the table structure and also produce compilable syntax. This is most visible for weaker VLMs such as LLaVA variants, where \latex{} usability is near zero despite non-trivial \html{} or \markdown{} usability. Even strong models show a consistent drop on \latex{} targets, with topology and content scores often lower than corresponding \html{} or \markdown{} outputs. Among open models, Qwen3-VL-8B is highly competitive and often exceeds larger models on both topology and content, while TableLLaVA remains much weaker. Overall, modern VLMs can recover table topology reliably from images, but exact content preservation and syntactically usable \latex{} generation remain challenging.

\section{Conclusion and Future Work}
\label{sec:conclusion}

We introduce \ds{}, a controlled multimodal table benchmark for studying representation effects in table understanding. \ds{} provides aligned \html{}, \markdown{}, and \latex{} representations with corresponding rendered images, category and difficulty tags, and a 700-sample balanced evaluation set drawn from five TableQA sources. This setup supports matched evaluation of QA, SUC, and SR across text and image inputs while keeping table content fixed. Our results show that table representation strongly affects model behavior. Structured text often outperforms rendered images, especially for structure-sensitive tasks, while \html{} is often the most robust format for text inputs and \latex{} remains challenging. In QA, verification questions are easier than Multi-Item Lookup and Aggregation/Arithmetic. In SUC, models handle column-oriented subtasks better than row and cell indexing, and header/indexing conventions can substantially shift answers. In SR, strong VLMs recover broad table layout reliably, but exact cell content and syntactically usable \latex{} generation remain difficult. Overall, scaling alone does not guarantee better table understanding. Future work can extend \ds{} to more realistic settings, including PDFs with noisy layouts and multiple tables, and explore methods for improving cross-format consistency and representation-aware decoding.

\clearpage
\section*{Limitations}
\label{sec:limitations}

TABVERSE targets controlled cross-format and cross-modality evaluation, so we make a few scope choices.
We build the benchmark from five established English TableQA sources (\textsc{FEVEROUS}, \textsc{HybridQA}, \textsc{SQA}, \textsc{TabFact}, \textsc{WIKITQ}) and focus on single-table questions answerable from the table alone; extending coverage to additional languages, scripts, and document-level settings is a natural next step.
For VLM-Image, we render table images from clean markup under a standardized layout. This keeps the image inputs aligned across \html{}, \markdown{}, and \latex{} and helps us isolate representation effects, but it does not capture noise from scanned or photographed documents.

\section*{Ethical Statement and Broad Impact}

\ds{} is constructed from publicly available datasets: \feverous{}, \hybridqa{}, \sqa{}, \tabfact{}, and \textsc{WIKITQ}. 
The benchmark is intended for academic research on multimodal and cross-format table understanding. 
We use released table-question pairs and do not add any private or proprietary data. 
Because \ds{} builds on existing datasets, it may inherit domain, linguistic, or annotation biases present in the original sources. 
\ds{} aims to support transparent and reproducible evaluation of LLMs and VLMs, and is not designed for commercial deployment or decision-making in sensitive domains.

\bibliography{custom}

\appendix
\clearpage
\section{Implementation and Evaluation Details}
\label{sec:models_evaluated}

\subsection{Model list}
We evaluate a set of text-only language models (LLMs) and vision--language models (VLMs), including both general-purpose and table-oriented models.
The evaluated models are:

\paragraph{Language Models (text-only):}
\emph{Qwen2.5-7B-Instruct},
\emph{Qwen3-30B-A3B-Instruct},
\emph{TableGPT2-7B},
\emph{TAMA-QWen3}.

\paragraph{Vision--Language Models:}
\emph{SmolVLM2-2.2B-Instruct},
\emph{Gemma-3-12B-IT},
\emph{Gemma-3-27B-IT},
\emph{InternVL3.5-14B},
\emph{InternVL3.5-30B-A3B},
\emph{Qwen3-VL-8B-Instruct},
\emph{Qwen3-VL-30B-A3B-Instruct},
\emph{Ministral-3-14B-Instruct},
\emph{LLaVA-1.6-7B}$^\ast$,
\emph{LLaVA-1.6-13B}$^\ast$,
\emph{TableLLaVA-v1.5-7B}$^\ast$,
\emph{GPT-5.2},
\emph{Gemini-3-Flash-Preview}.

\subsection{Pipelines and configurations}
We evaluate:
(i)~\emph{LLM} (structural-text) for text-only models,
(ii)~\emph{VLM-Image} (rendered table images) for VLMs, and
(iii)~\emph{VLM-Text} (structural-text) for VLMs when the model interface supports text-only operation.
For starred models ($^\ast$), we report results only for the configurations that are supported reliably by the model interface and context window.

\subsection{Context length and coverage}
Structural-text evaluation includes the full table markup in the prompt and can require long context.
When a prompt exceeds a model's supported context length (or fails to run reliably), we mark that instance as out of coverage for that model--configuration and compute metrics over the remaining evaluable instances.
We report per-model coverage statistics alongside results. 
We ran open-weight model experiments on NVIDIA A100 SXM GPUs, using GPU execution for inference and supervised fine-tuning. Closed-model experiments were conducted through the OpenAI and Google Gemini APIs. 

\subsection{Decoding and post-processing}
All evaluations use zero-shot prompting with greedy decoding (\texttt{temperature}=0, \texttt{top\_p}=1) and task-specific output limits.
We apply minimal normalization for scoring consistency (e.g., stripping boilerplate prefixes, whitespace normalization, and simple label extraction for verification tasks).
Prompt templates and exact decoding limits are listed in Appendix~\ref{sec:prompts}.

\subsection{Additional Metric Details}
\label{app:metric_details}

For QA, we report exact-match accuracy after light normalization. Single-answer questions require an exact normalized match, while multi-item lookup questions require the normalized predicted set to match the gold set. We also report a relaxed QA metric that counts a prediction as correct when the gold answer appears as a complete normalized span. Numeric answers are compared after extracting and normalizing numeric strings. For SUC, exact match is the primary metric, with Field Accuracy and Relaxed Accuracy reported as diagnostics for structured-answer errors. Field Accuracy compares pipe-separated fields position-wise, giving partial credit when only some fields are correct:

\begin{equation}
\mathrm{FieldAcc}_i =
\frac{1}{K_i}\sum_{k=1}^{K_i}\mathbb{1}[\hat{y}_{i,k}=y_{i,k}]
\label{eq:field_accuracy}
\end{equation}

where $K_i$ is the number of gold fields. 
Relaxed Accuracy measures how many gold fields appear somewhere in the model output:

\begin{equation}
\mathrm{RelaxedAcc}_i =
\frac{1}{K_i}\sum_{k=1}^{K_i}\mathbb{1}[y_{i,k}\in \hat{y}_i]
\label{eq:relaxed_accuracy}
\end{equation}

These diagnostics distinguish fully incorrect predictions from outputs containing the correct values in the wrong format or order. For SR, output usability is evaluated separately by target format: HTML must yield a parsable table, Markdown must render to a recoverable table, and LaTeX must compile successfully with a fixed wrapper. No repair is applied to malformed outputs.

\section{Dataset Details}
\label{app:dataset-details}

\subsection{Source splits, filtering, and overlap checks}
We construct the raw pool from the official held-out split for each source dataset (test when available; otherwise dev) and filter to retain single-table, table-grounded instances. For datasets with mixed table/passage supervision (e.g., \hybridqa{} and \feverous{}), we drop instances whose gold evidence requires non-table context or multiple tables. We also check for overlaps against the corresponding training splits and remove duplicated question--table pairs when detected. Table~\ref{tab:tagged-pool-stats} summarizes the resulting tagged pool.

\begin{table}[t]
\centering
\small
\setlength{\tabcolsep}{4pt}
\begin{tabular}{lccc}
\toprule
\textbf{Dataset} &
\textbf{Split} &
\textbf{\# Questions} &
\textbf{\# Tables} \\
\midrule
\feverous{}      & dev           & 794  & 525  \\
\hybridqa{}      & dev           & 1608 & 1608 \\
\tabfact{}       & test          & 1695 & 1695 \\
\sqa{}           & test          & 1000 & 185  \\
\textsc{wikitq}  & unseen tables & 1000 & 421  \\
\midrule
\textbf{Total}   & --            & 6097 & 4434 \\
\bottomrule
\end{tabular}
\caption{
Composition of the tagged pool after filtering and normalization. \#Questions denotes retained question instances, and \#Tables denotes unique underlying tables.
}
\label{tab:tagged-pool-stats}
\end{table}

\subsection{Tagged pool normalization}
The source datasets use different schemas for tables and supervision.
We normalize each example into a common JSON format (table content, question, answer/label, and metadata) and assign a stable \texttt{table\_id} to each unique table so that multiple questions can reference the same underlying table.

\subsection{Format conversion to \html{}, \markdown{}, and \latex{}}
We convert each table into three structural formats.
We adapt conversion utilities released with \citet{sui2024table} and extend them to (i) target our held-out splits, (ii) enforce consistent row/column ordering across formats, and (iii) produce syntactically valid outputs under dataset-specific artifacts (e.g., missing values and special characters).
We generate \html{} markup with standard \texttt{<table>}/\texttt{<tr>}/\texttt{<th>}/\texttt{<td>} tags, \markdown{} tables with pipe-delimited syntax, and compilable \latex{} \texttt{tabular} code with appropriate escaping.

\subsection{Rendering pipelines}
We render a table image from each structural representation under a standardized layout (font size, padding, width).
We render \html{} tables in a controlled browser environment, convert \markdown{} to \html{} before rendering, and compile \latex{} to PDF before converting to PNG.
Because we render images from the generated markup, the rendered images are aligned with the textual tables by construction.

\subsection{Question category taxonomy}
\label{app:qcat-taxonomy}
We use seven question categories for analysis and stratification (the same set used in the Results section).
For each category, we provide a short definition and one example prompt in Appendix~\ref{sec:prompts}.

\section{Prompts}
\label{sec:prompts}

We use a dedicated prompt to assign each table--question pair to one of the predefined reasoning categories described in Section~\ref{sec:dataset}. The full classification prompt is shown in Figure~\ref{fig:qcat-prompts}.

\vspace{1em}
\noindent\begin{tcolorbox}[promptbox, title=Prompt: Question Category Classification]
\small

\textbf{System Prompt}

You are an expert at classifying table-question pairs into Question Categories.  
You must follow the rules exactly and output exactly ONE line in the required format.

\vspace{0.5em}

\textbf{Input}

\textbf{TABLE (as JSON with header + rows):}

\texttt{\{table\_json\}}

\textbf{QUESTION}

\texttt{\{query\}}

\vspace{0.6em}

\textbf{TASK}

Assign exactly ONE Question Category using the definitions below.

\vspace{0.6em}

\textbf{CRITICAL RULE (TABLE-REQUIREDNESS)}

A question can be assigned a predefined Question Category  
ONLY IF the table is REQUIRED to produce the final answer.

External knowledge policy:

You MAY use external knowledge to:
\begin{itemize}[leftmargin=1.3em]
\item Map a description or condition to an entity present in the table
(e.g., ``country with population 67.02 million'' → ``France'')
\item Resolve aliases, real-world facts, or descriptive constraints 
ONLY for the purpose of identifying the correct row(s)
\end{itemize}

You MUST NOT:
\begin{itemize}[leftmargin=1.3em]
\item Output a final answer that does not come directly from table cells
\item Compute the final answer using non-table facts
\item Label as answerable if the table is not necessary
\end{itemize}

If the final answer can be obtained without using the table at all, output:

\texttt{None of the above — Not table-required}

If the required answer value does not exist in or cannot be computed from the table cells, output:

\texttt{None of the above — Missing answer attribute in table}

\vspace{0.6em}

\textbf{PROCEDURE}

\textbf{Step A — Table-Requiredness Check}

\begin{enumerate}[leftmargin=1.3em]
\item Identify what the question asks for:
\begin{itemize}
\item single value
\item list/set
\item number
\item yes/no
\end{itemize}

\item Identify which table cell(s) must be read or aggregated to produce the final answer.

\item Verify that the final answer is directly read from or computed using ONLY table cells.

\item If the table is not needed to produce the final answer →  
None of the above — Not table-required

\item If the answer value is not present or cannot be computed from table cells →  
None of the above — Missing answer attribute in table
\end{enumerate}

Otherwise, proceed to classification.

\vspace{0.6em}

\textbf{QUESTION CATEGORIES (Choose EXACTLY ONE)}

\begin{itemize}[leftmargin=1.3em]

\item \textbf{Simple Lookup}  
Identify ONE row and read ONE cell.  
(No filtering beyond locating that row.)

\item \textbf{Conditional Lookup}  
Apply one or more conditions to select row(s),  
then read ONE resulting value.

\item \textbf{Multi-Item Lookup}  
Return multiple values/rows from the table (a list/set).

\item \textbf{Aggregation / Counting / Arithmetic}  
Compute a number from table values  
(count/sum/avg/difference/ratio/percent/etc.).

\item \textbf{Comparison \& Extremum}  
Choose max/min/earliest/latest by comparing table values.

\item \textbf{Single-step Binary Verification}  
Verify one statement directly using the table.

\item \textbf{Multi-hop Binary Verification}  
Verify a statement requiring multiple reasoning steps  
across the table.

\end{itemize}

\vspace{0.6em}

\textbf{OUTPUT FORMAT (STRICT)}

Output exactly ONE line and nothing else:

\texttt{Question Category: <CATEGORY NAME>}

OR (if not answerable under rules):

\texttt{Question Category: None of the above}

\end{tcolorbox}
\captionof{figure}{Prompt used to classify table-question pairs into structured reasoning categories.}
\label{fig:qcat-prompts}

\vspace{1em}

For QA evaluation, we use a minimal answer-only prompt and a separate binary-verification variant for yes/no statements. The prompts are shown in Figure~\ref{fig:tableqa-prompts}.

\vspace{1em}
\noindent\begin{tcolorbox}[promptbox,title=Prompt: Table Question Answering]
\small

\textbf{General QA Prompt}

Look at the given table and answer the following question directly. 
Do not include introductions, explanations, or extra text. 
Provide only the exact, precise final answer.

\texttt{\{query\}}

\vspace{0.8em}

\textbf{Binary Verification Prompt}

Look at the given table and answer the following question with only a single digit: 
1 if the statement is true, 0 if the statement is false. 
Do not include any explanations or extra text.

\texttt{\{query\}}

\end{tcolorbox}
\captionof{figure}{Prompts used for table-based QA and binary verification tasks.}
\label{fig:tableqa-prompts}

\vspace{1em}

For Structure Reconstruction (SR), models are instructed to generate a complete table representation in the requested target format. The reconstruction prompts are shown in Figure~\ref{fig:gen-prompts}.

\vspace{1em}

\noindent\begin{tcolorbox}[promptbox,title=Prompt: Table Structure Generation]
\small

\textbf{HTML Generation Prompt}

Generate the complete HTML code that exactly represents this image. Provide only the code without any explanations.

\vspace{0.5em}

\textbf{LaTeX Generation Prompt}

Generate the complete LaTeX code that exactly represents this image. Provide only the code without any explanations.

\vspace{0.5em}

\textbf{Markdown Generation Prompt}

Generate the complete Markdown code that exactly represents this image. Provide only the code without any explanations.

\end{tcolorbox}
\captionof{figure}{Prompts used for generating structured table representations from images in HTML, LaTeX, and Markdown formats.}
\label{fig:gen-prompts}

\vspace{0.8em}

For Structure Understanding and Cell-level Tasks (SUC), we use task-specific prompts covering boundary detection, table size estimation, coordinate lookup, and row/column retrieval. The complete prompt set is shown in Figure~\ref{fig:suc-prompts}.

\vspace{1em}

\noindent\begin{tcolorbox}[promptbox,title=Prompt: Table Structure and Cell-Level Tasks]
\small

\textbf{table\_partition}

What is the **first cell value** (not including headers) of the given table? What is the **last cell value** (not including headers) of the given table? Answer questions one by one and use | to split the answer. Answer the question without having any introduction or explanations.

\vspace{0.6em}

\textbf{table\_first\_cell}

What is the **first cell value** (not including headers) of the given table? Answer the question without having any introduction or explanations.

\vspace{0.6em}

\textbf{table\_last\_cell}

What is the **last cell value** (not including headers) of the given table? Answer the question without having any introduction or explanations.

\vspace{0.6em}

\textbf{size\_detection}

How many rows in the table? How many columns in the table? Answer the questions one by one and use | to split the answer. Answer the question without having any introduction or explanations.

\vspace{0.6em}

\textbf{number\_of\_rows}

How many rows in the table? Answer the question without having any introduction or explanations.

\vspace{0.6em}

\textbf{number\_of\_columns}

How many columns in the table? Answer the question without having any introduction or explanations.

\vspace{0.6em}

\textbf{cell\_lookup}

Row/column indices start at 0 (top-left is 0|0). What is the position of the cell value \texttt{\{cell\_value\}}? Use row index and column index to answer. Use | to split the answer. Answer the question without having any introduction or explanations.

\vspace{0.6em}

\textbf{reverse\_lookup}

Row/column indices start at 0 (top-left is 0|0). What is the cell value of row index \texttt{\{reverse\_lookup\_row\}}, column index \texttt{\{reverse\_lookup\_col\}} ? Only output the cell value without other information. Answer the question without having any introduction or explanations.

\vspace{0.6em}

\textbf{column\_retrieval}

Row/column indices start at 0 (top-left is 0|0). What is the column name with the index \texttt{\{column\_idx\}} of the given table image? Only give the column name without any explanation. Answer the question without having any introduction or explanations.

\vspace{0.6em}

\textbf{row\_retrieval}

Row/column indices start at 0 (top-left is 0|0). What are the cell values of the \texttt{\{row\_idx\}} row in following table? Only list the cell values one by one using | to split the answers. Answer the question without having any introduction or explanations.

\end{tcolorbox}
\captionof{figure}{Prompts used for evaluating structural understanding}
\label{fig:suc-prompts}

\vspace{0.8em}

\section{Result Discussion}
\subsection{TaskQA: Additional Analyses}
\label{app:taskqa}

This appendix provides additional TaskQA analyses that complement the main results. We examine modality gaps, strict-versus-relaxed matching, question-category performance, and Easy/Hard difficulty breakdowns to better understand model behavior across table formats and reasoning types.

\begin{figure}[!h]
  \centering
  \includegraphics[width=0.95\linewidth]{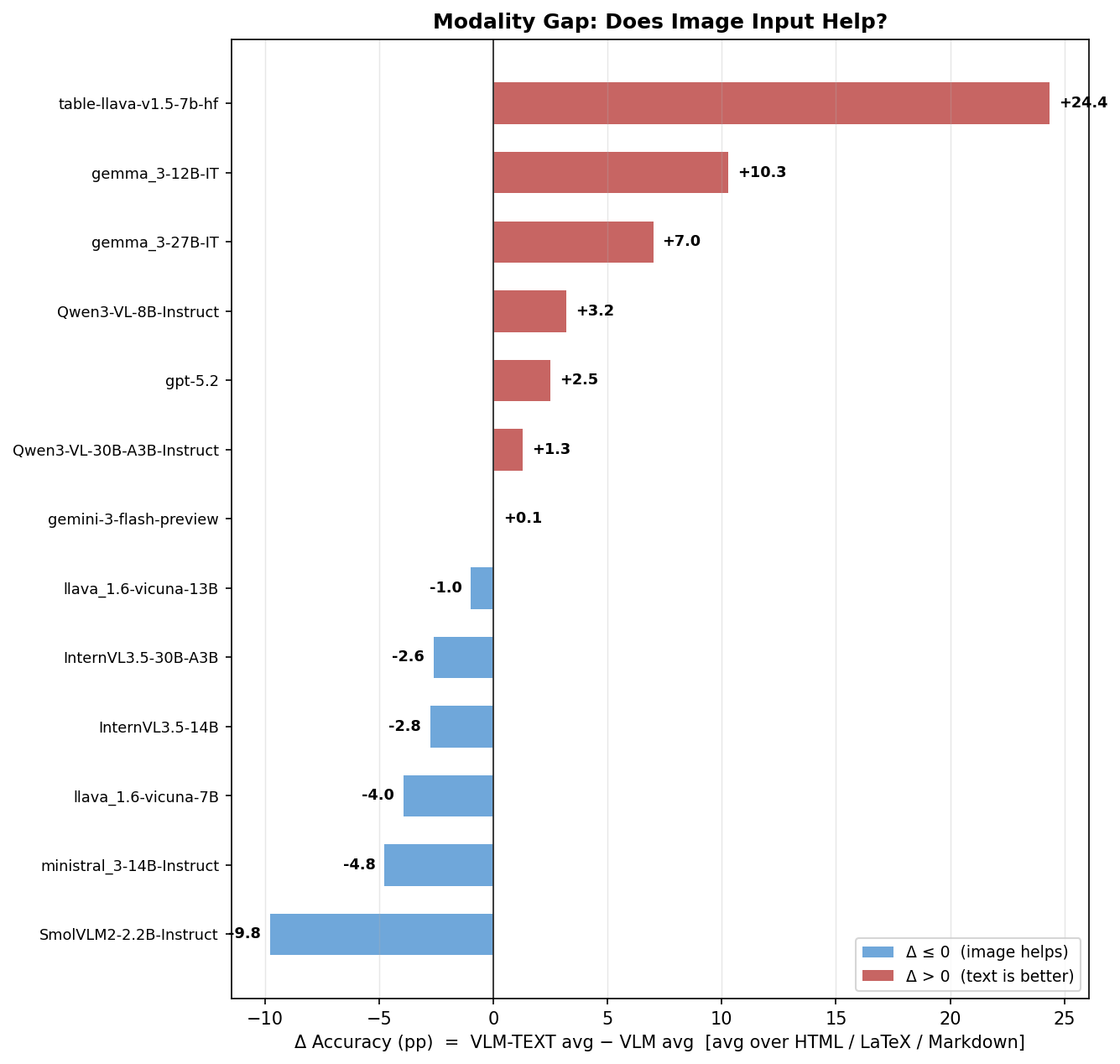}
  \caption{\textbf{TaskQA modality gap.} $\Delta$ accuracy (pp) $=$ VLM-Text avg $-$ VLM-Image avg, averaged over \html{}/LaTeX/\markdown{}. Negative means images help more than text.}
  \label{fig:task_modality_gap}
\end{figure}

\paragraph{Modality gap:} Figure~\ref{fig:task_modality_gap} reports 
$\Delta = \text{VLM-Text avg} - \text{VLM-Image avg}$,
averaged over \html{}, LaTeX, and \markdown{}. Positive values mean that structured table text helps more than rendered images, while negative values mean that rendered images help more. The direction of the gap depends on the model family. Gemma-3 shifts toward structured text, with gains of +10.3 points for the 12B model and +7.0 points for the 27B model. In contrast, SmolVLM2 and InternVL3.5 shift toward rendered images. Gemini-3-Flash-Preview stays near zero, which is consistent with its stable performance across modalities in Table~\ref{tab:taskqa}. These results suggest that modality preferences are model-dependent rather than a universal property of VLM-based table reasoning.

\paragraph{Strict vs relaxed matching:} Table~\ref{tab:taskqa_em_relaxed_all} reports both strict exact-match and relaxed accuracy. Relaxed accuracy counts a prediction as correct when the normalized gold answer appears as a complete answer span within a longer response. This diagnostic separates answer retrieval from answer-only formatting rather than replacing strict EM. For example, Qwen3-VL-8B-IT remains stronger under strict EM, while Qwen3-VL-30B-A3B-IT improves substantially under relaxed matching, especially on structured table-text inputs, suggesting that some EM errors stem from verbose formatting rather than answer retrieval failures.

\begin{table*}[t]
\centering
\small
\setlength{\tabcolsep}{4pt}
\renewcommand{\arraystretch}{1}

\definecolor{headerbg}{HTML}{FCE5CD}
\definecolor{openbg}{HTML}{F4CCCC}
\definecolor{vlmbg}{HTML}{D9EAD3}
\definecolor{propbg}{HTML}{D9D2E9}
\definecolor{top}{HTML}{D9EAF7}

\begin{tabular}{l cc cc cc cc cc cc}
\toprule
& \multicolumn{6}{c}{\cellcolor{top}\textbf{Table Image Render}} 
& \multicolumn{6}{c}{\cellcolor{top}\textbf{Table Text Format}} \\
\cmidrule(lr){2-7}\cmidrule(lr){8-13}

\rowcolor{headerbg}
\textbf{Model} 
& \multicolumn{2}{c}{\textbf{\html{}}}
& \multicolumn{2}{c}{\textbf{LaTeX}}
& \multicolumn{2}{c}{\textbf{\markdown{}}}
& \multicolumn{2}{c}{\textbf{\html{}}}
& \multicolumn{2}{c}{\textbf{LaTeX}}
& \multicolumn{2}{c}{\textbf{\markdown{}}} \\

\rowcolor{headerbg}
& \textbf{EM} & \textbf{Rel.}
& \textbf{EM} & \textbf{Rel.}
& \textbf{EM} & \textbf{Rel.}
& \textbf{EM} & \textbf{Rel.}
& \textbf{EM} & \textbf{Rel.}
& \textbf{EM} & \textbf{Rel.} \\
\midrule

\rowcolor{openbg}
\multicolumn{13}{l}{\textbf{Language Models (text-only)}} \\

Qwen2.5-IT
& -- & --
& -- & --
& -- & --
& 44.57 & 50.29
& 42.71 & 48.43
& 45.43 & 51.43 \\

Qwen3-IT
& -- & --
& -- & --
& -- & --
& 51.14 & 61.43
& 48.43 & 60.86
& 46.57 & 62.57 \\

TableGPT2
& -- & --
& -- & --
& -- & --
& 44.43 & 57.29
& 41.57 & 56.00
& 42.14 & 58.57 \\

TAMA-QWen3
& -- & --
& -- & --
& -- & --
& 18.29 & 52.00
& 19.14 & 53.71
& 20.71 & 52.71 \\

\addlinespace[3pt]
\rowcolor{vlmbg}
\multicolumn{13}{l}{\textbf{Vision-Language Models}} \\

SmolVLM2-IT
& 29.71 & 31.57
& 28.71 & 30.71
& 25.86 & 30.00
& 21.57 & 39.80
& 17.63 & 34.39
& 15.75 & 33.67 \\

Gemma-3-IT 12B
& 38.86 & 41.86
& 39.57 & 43.29
& 38.57 & 43.00
& 50.29 & 55.57
& 49.00 & 54.14
& 48.57 & 53.71 \\

Gemma-3-IT 27B
& 46.14 & 50.29
& 45.29 & 48.57
& 45.43 & 49.43
& 53.43 & 59.14
& 51.29 & 56.57
& 53.14 & 58.86 \\

InternVL3.5 14B
& \underline{48.57} & 52.57
& 48.14 & 52.43
& 48.00 & 52.57
& \underline{47.14} & 55.86
& \underline{47.29} & 54.29
& 44.86 & 55.14 \\

InternVL3.5 30B-A3B
& 47.86 & \underline{54.29}
& \underline{50.00} & \underline{55.43}
& \underline{48.29} & \underline{54.71}
& 45.86 & \underline{57.57}
& 45.71 & \underline{56.14}
& \underline{47.00} & \underline{61.00} \\

Qwen3-VL-IT 8B
& \underline{50.29} & \underline{55.43}
& \underline{49.29} & \underline{54.71}
& \underline{49.71} & \underline{55.14}
& \underline{53.43} & 59.14
& \underline{52.14} & 57.43
& \underline{53.29} & 59.14 \\

Qwen3-VL-IT 30B-A3B
& 41.14 & 53.29
& 42.14 & 54.29
& 41.43 & 53.29
& 45.29 & \underline{62.86}
& 43.57 & \underline{60.86}
& 39.71 & \underline{63.00} \\

Ministral-3-IT
& 44.43 & 55.29
& 39.14 & 49.29
& 42.71 & 53.00
& 40.00 & 50.86
& 35.43 & 46.86
& 36.57 & 46.57 \\

LLaVA-1.6 7B
& \underline{31.86} & \underline{35.71}
& \underline{31.43} & \underline{35.29}
& \underline{32.00} & \underline{35.71}
& \underline{27.37} & \underline{43.91}
& \underline{29.55} & 41.05
& \underline{26.50} & 42.17 \\

LLaVA-1.6 13B
& 25.14 & 27.57
& 23.71 & 26.43
& 25.00 & 27.43
& 23.91 & 42.56
& 22.13 & \underline{43.67}
& 24.89 & \underline{44.07} \\

\addlinespace[3pt]
\rowcolor{vlmbg}
\multicolumn{13}{l}{\textbf{Table-specialised Vision-Language Models}} \\

TableLLaVA-v1.5
& 1.29 & 20.86
& 1.00 & 19.71
& 4.00 & 23.71
& 23.61 & 36.24
& 27.37 & 37.41
& 28.40 & 38.80 \\

\addlinespace[3pt]
\rowcolor{propbg}
\multicolumn{13}{l}{\textbf{Proprietary Models}} \\

GPT-5.2
& 54.57 & 61.71
& 54.52 & 61.12
& 56.14 & 63.29
& 57.43 & 66.00
& 57.29 & 65.29
& 58.00 & 66.57 \\

Gemini-3-Flash-Preview
& 65.43 & 72.00
& 65.14 & 71.16
& 65.43 & 71.57
& 65.71 & 72.29
& 65.00 & 71.86
& 65.43 & 72.86 \\

\bottomrule
\end{tabular}

\caption{
\textbf{TaskQA strict and relaxed matching diagnostic.}
Exact-match accuracy (EM) and relaxed accuracy (Rel.) are reported across \html{}, LaTeX, and \markdown{} inputs.
Relaxed accuracy counts a prediction as correct when the normalized gold answer appears as a complete answer span inside a longer response.
Underlined values indicate the highest score within each selected model-variant group and column.
This diagnostic does not replace EM; it highlights cases where models retrieve the correct answer but fail the answer-only format required by strict exact match.
}
\label{tab:taskqa_em_relaxed_all}
\end{table*}

\begin{figure}[!h]
  \centering
  \includegraphics[width=0.90\linewidth]{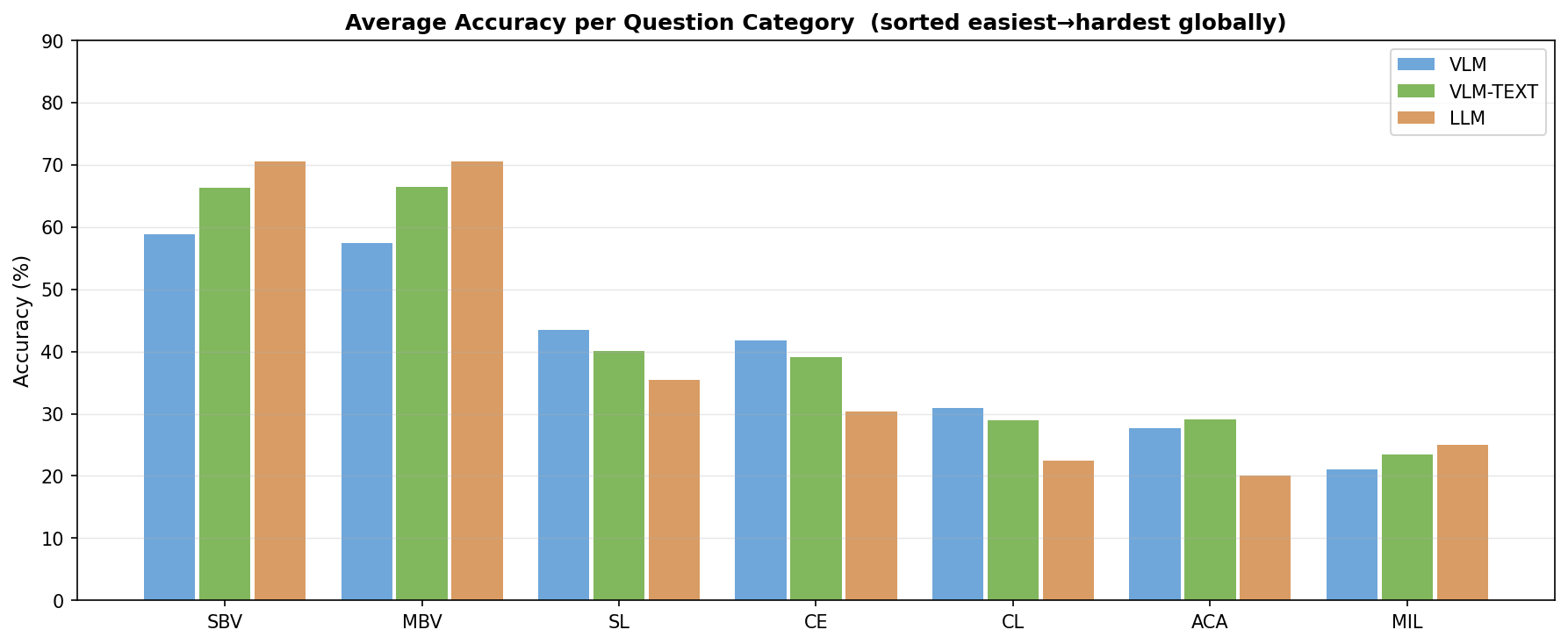}
  \caption{\textbf{TaskQA category averages.} Accuracy by question category, averaged over models and formats, shown per pipeline (VLM-Image / VLM-Text / LLM-Text).}
  \label{fig:task_category_bar}
\end{figure}

\begin{figure}[!h]
    \centering
    \includegraphics[width=0.75\linewidth]{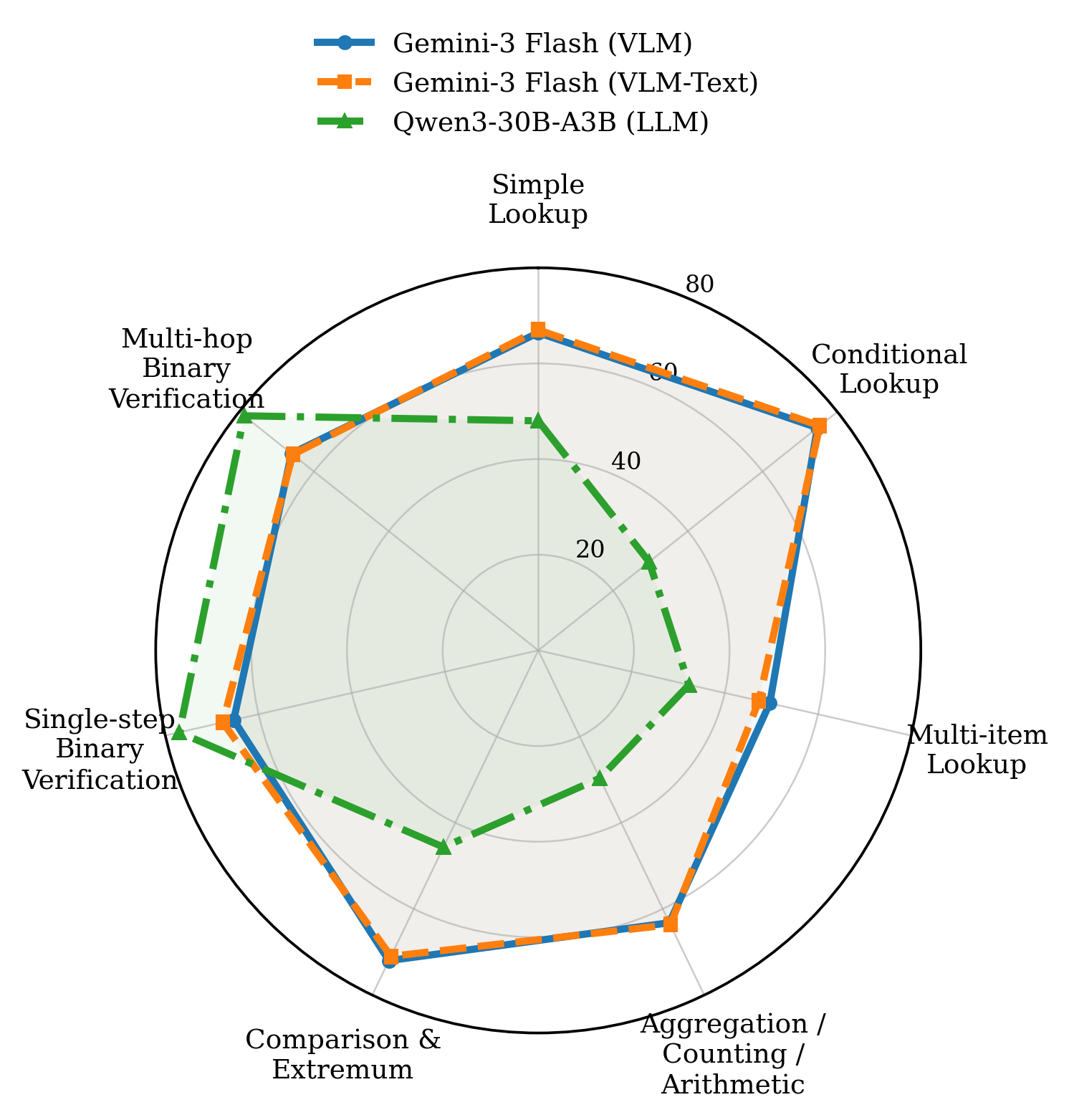}
\caption{
Category-wise Task QA accuracy for the strongest model from each pipeline.
}
\label{fig:task-category-radar}
\end{figure}

\paragraph{Question categories:} Figure~\ref{fig:task_category_bar} reports accuracy by question category, averaged over models and formats.
Verification-style questions are easier across pipelines, while multi-item lookup and aggregation/counting questions are more difficult. This suggests that models handle binary or localized evidence better than questions requiring multiple retrieved items, counting, or arithmetic operations. The category averages in Figure~\ref{fig:task_category_bar} mask differences among the strongest models. Figure~\ref{fig:task-category-radar} shows that Gemini-3-Flash performs consistently well across most categories, while Qwen3-30B-A3B is strongest on multi-hop binary verification. The two Gemini-3-Flash variants perform similarly, indicating limited modality effects. In contrast, multi-item lookup and aggregation/counting remain among the weakest categories across pipelines, highlighting the difficulty of retrieval and composition.

\begin{figure*}[t]
  \centering
  \includegraphics[width=0.95\linewidth]{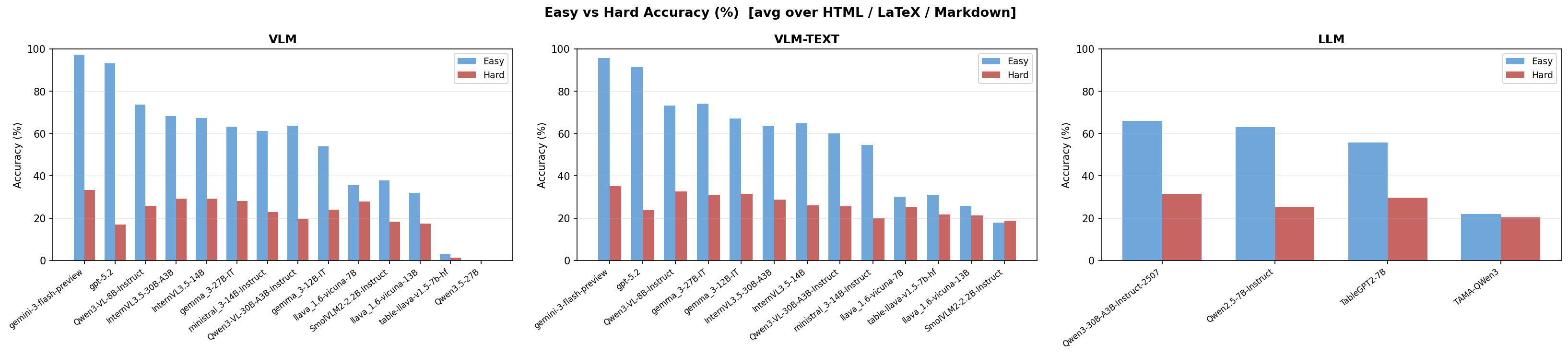}
  \caption{\textbf{TaskQA Easy vs Hard.} Easy and Hard exact-match accuracy per model and pipeline (VLM-Image / VLM-Text / LLM-Text), averaged over \html{}/LaTeX/\markdown{}.}
  \label{fig:task_difficulty}
\end{figure*}

\paragraph{Easy vs Hard split:}
Figure~\ref{fig:task_difficulty} reports Easy and Hard accuracy for each model and pipeline, averaged over \html{}, LaTeX, and \markdown{}. All pipelines show a clear drop from Easy to Hard questions, confirming that the difficulty annotation captures increased reasoning or evidence-composition demands. The gap is particularly large for the strongest VLMs, whose Easy accuracy often exceeds 80--90\% while Hard accuracy remains below 40\%, indicating that multi-step reasoning remains a major challenge despite strong overall performance. We use this split as a diagnostic view of difficulty rather than a definition of reasoning complexity.

\paragraph{Evaluation notes:} TaskQA is scored with strict exact-match accuracy in the main results. We apply the same normalization and post-processing to all models. Strict EM is intentionally conservative: answers with extra explanatory text are counted as incorrect even when they contain the gold answer. For this reason, Table~\ref{tab:taskqa_em_relaxed_all} provides a diagnostic relaxed-matching view. 

Models marked with $^*$ have shorter context windows; when they fail to return an answer on large-table cases, the output is counted as incorrect under the same scoring rule. This ensures consistent evaluation across architectures and context lengths.

\begin{figure}[!h]
\centering
\includegraphics[
    width=\columnwidth,
    trim=3cm 4cm 4cm 3cm,
    clip
]{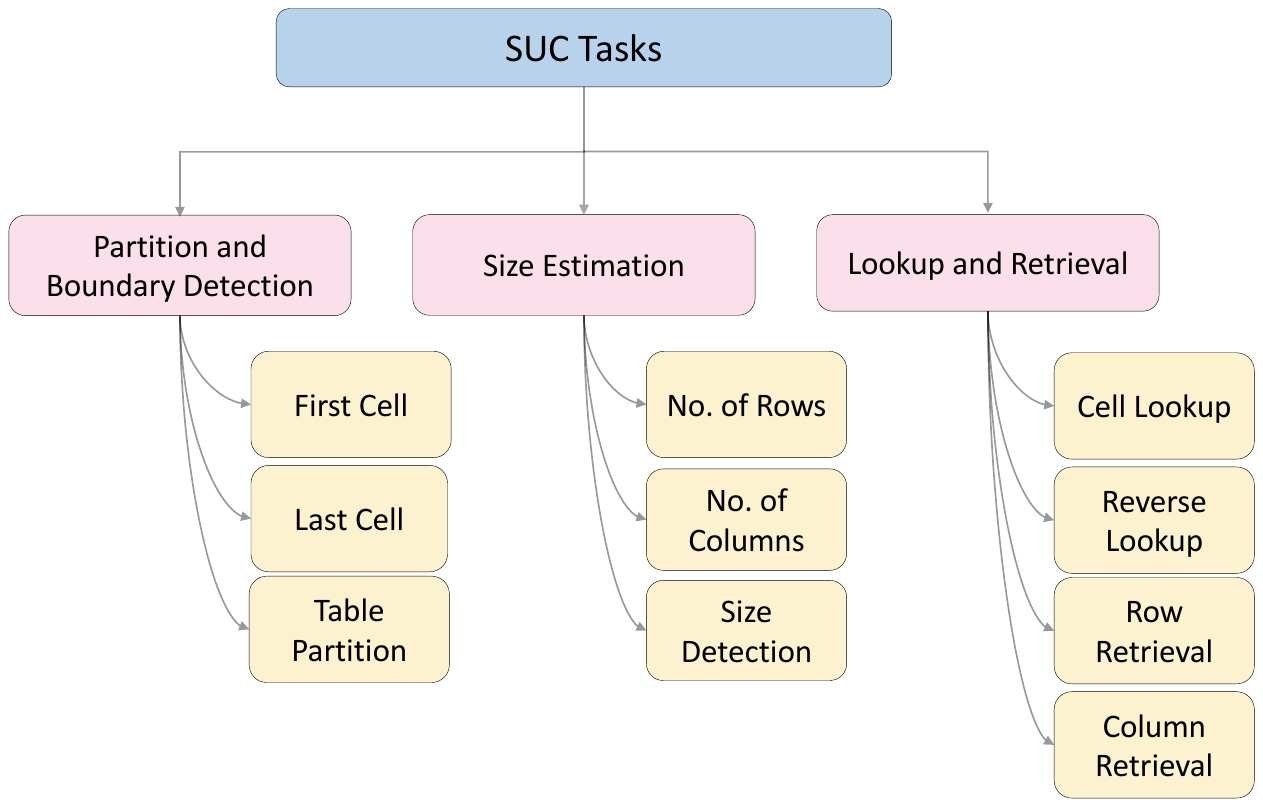}
\caption{Taxonomy of SUC tasks. Ten tasks are grouped into partitioning, size estimation, lookup, and retrieval.}
\label{fig:task_taxonomy}
\end{figure}

\begin{figure}[!h]
  \centering
  \includegraphics[width=0.95\linewidth]{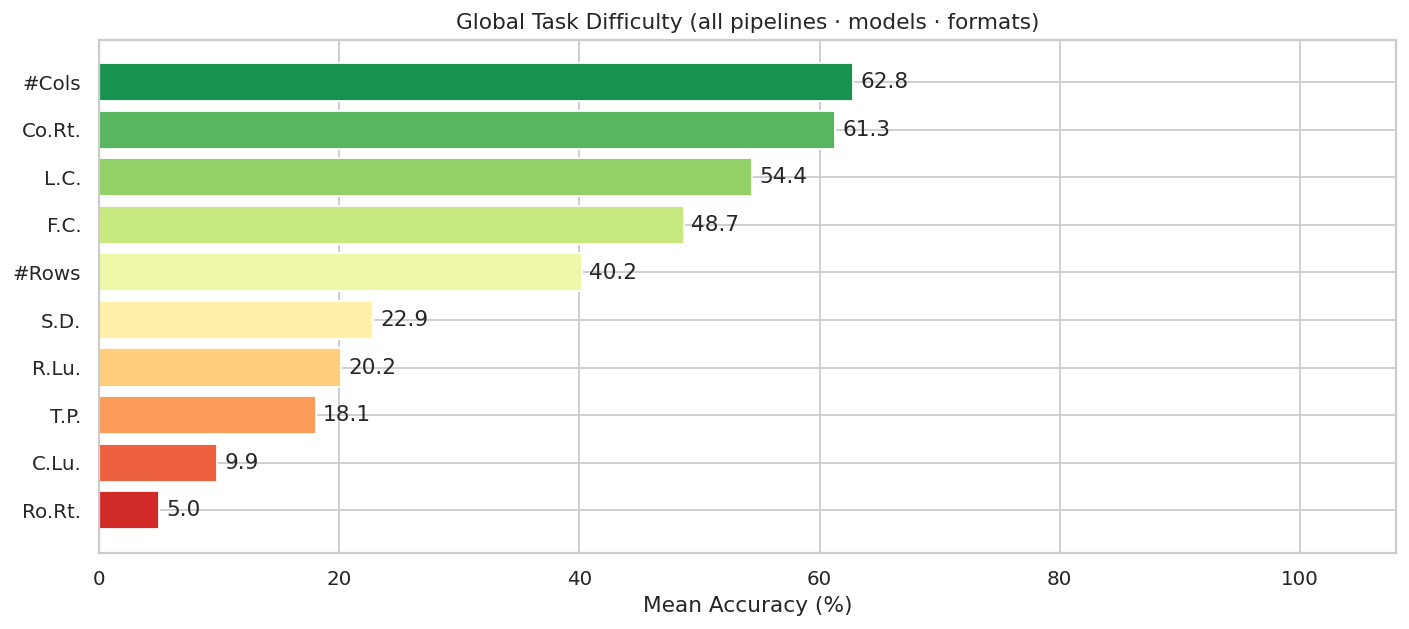}
  \caption{\textbf{SUC task difficulty.} We average exact-match accuracy over all models, pipelines, and formats for each SUC subtask. Higher is better.}
  \label{fig:suc_task_difficulty}
\end{figure}

\subsection{Structural Understanding Capability: Additional Analyses}
\label{app:suc}

This appendix provides additional SUC analyses that support Section~\ref{sec:suc}. Figure~\ref{fig:task_taxonomy} summarizes the SUC task taxonomy. We include subtask-level difficulty, pipeline comparisons, diagnostic metrics, prompt-sensitivity results, format effects, and evaluation notes.

\paragraph{Subtask difficulty:}
Figure~\ref{fig:suc_task_difficulty} shows that SUC difficulty is highly subtask-dependent.
Column counting and column retrieval are consistently among the easiest subtasks across pipelines.
In contrast, row retrieval, cell lookup, table partitioning, and size detection remain difficult.
This pattern shows that models are better at identifying global column structure than at recovering precise row-level or coordinate-based structure.

\begin{table*}[t]
\centering
\small
\setlength{\tabcolsep}{3.2pt}
\renewcommand{\arraystretch}{0.72}

\definecolor{headerbg}{HTML}{FCE5CD}
\definecolor{openbg}{HTML}{F4CCCC}
\definecolor{tablebg}{HTML}{D9EAD3}

\begin{tabular}{l c c c c c c c c c c c c}
\toprule
\rowcolor{headerbg}
\textbf{Models} & \textbf{Formats} &
\textbf{T.P.} & \textbf{F.C.} & \textbf{L.C.} &
\textbf{S.D.} & \textbf{\# Rows} & \textbf{\# Cols} &
\textbf{C.Lu.} & \textbf{R.Lu.} & \textbf{Co.Rt.} & \textbf{Ro.Rt.} &
\textbf{Overall} \\
\midrule

\rowcolor{openbg}
\multicolumn{13}{l}{\textbf{Open Models}} \\

\multirow{3}{*}{SmolVLM2-2.2B-IT}
 & \html{}     & 0.2 & 48.6 & 31.6 & 2.1 & 26.9 & 8.9 & 1.1 & 6.2 & 35.0 & 0.0 & 16.1 \\
 & LaTeX       & 0.0 & 49.4 & 34.5 & 0.2 & 18.9 & 8.4 & 1.1 & 7.0 & 30.4 & 0.0 & 15.0 \\
 & \markdown{} & 0.0 & 19.6 & 17.0 & 0.5 & 2.7 & 0.6 & 0.5 & 6.2 & 5.6 & 0.0 & 5.3 \\

\multirow{3}{*}{Gemma-3-12B-IT}
 & \html{}     & 10.3 & 37.5 & 45.8 & 13.7 & 30.0 & 35.0 & 10.3 & 14.5 & 56.3 & 2.9 & 25.6 \\
 & LaTeX       & 13.8 & 41.2 & 48.2 & 13.5 & 30.2 & 37.2 & 11.0 & 15.1 & 62.8 & 3.0 & 27.6 \\
 & \markdown{} & 10.7 & 40.1 & 46.7 & 18.1 & 32.8 & 42.0 & 14.1 & 15.4 & 66.8 & 2.4 & 28.9 \\

\multirow{3}{*}{InternVL3.5-30B-A3B}
 & \html{}     & 0.0 & 88.9 & 79.0 & 8.7 & 39.9 & 88.6 & 17.8 & 9.2 & 53.7 & 2.2 & 38.8 \\
 & LaTeX       & 0.0 & 89.2 & 80.6 & 12.6 & 44.5 & 83.5 & 21.5 & 15.9 & 69.8 & 4.0 & 42.1 \\
 & \markdown{} & 0.0 & 84.9 & 76.5 & 25.1 & 56.1 & 85.5 & 24.5 & 13.0 & 64.9 & 4.5 & 43.5 \\

\multirow{3}{*}{Qwen3-VL-8B-IT}
 & \html{}     & 23.8 & 93.3 & 80.9 & 8.1 & 25.6 & 89.3 & 27.0 & 26.6 & 90.1 & 1.0 & 46.6 \\
 & LaTeX       & 20.8 & 93.2 & 78.9 & 14.3 & 36.1 & 88.7 & 32.4 & 30.0 & 89.5 & 3.0 & 48.7 \\
 & \markdown{} & 24.6 & 87.8 & 78.4 & 14.6 & 31.8 & 90.1 & 28.0 & 34.2 & 89.8 & 2.2 & 48.2 \\

\multirow{3}{*}{LLaVA-1.6-7B}
 & \html{}     & 0.0 & 14.3 & 14.5 & 3.2 & 18.9 & 30.2 & 1.7 & 1.3 & 29.6 & 0.0 & 11.4 \\
 & LaTeX       & 0.3 & 13.8 & 18.1 & 1.3 & 20.5 & 23.5 & 1.9 & 2.9 & 28.3 & 0.0 & 11.1 \\
 & \markdown{} & 0.2 & 14.1 & 18.1 & 1.9 & 17.6 & 23.8 & 1.6 & 3.3 & 29.9 & 0.0 & 11.1 \\

\addlinespace[4pt]
\rowcolor{tablebg}
\multicolumn{13}{l}{\textbf{Table-specialised Models}} \\

\multirow{3}{*}{TableLLaVA-v1.5-7B}
 & \html{}     & 0.0 & 6.5 & 4.3 & 0.0 & 0.0 & 0.3 & 0.0 & 1.6 & 1.0 & 0.0 & 1.4 \\
 & LaTeX       & 0.0 & 2.2 & 4.3 & 0.0 & 2.1 & 2.9 & 0.0 & 1.7 & 3.7 & 0.0 & 1.7 \\
 & \markdown{} & 0.0 & 6.7 & 9.2 & 0.0 & 13.0 & 14.5 & 0.0 & 2.7 & 9.2 & 0.0 & 5.5 \\

\bottomrule
\end{tabular}

\caption{
\textbf{Additional SUC results for the VLM-Image pipeline.}
Exact-match accuracy (\%) across ten structure-oriented subtasks.
Models receive rendered table images derived from \html{}, LaTeX, and \markdown{} sources.
}
\label{tab:suc_vlm_image_appendix}
\end{table*}

\begin{figure*}[t]
  \centering
  \includegraphics[width=0.95\linewidth]{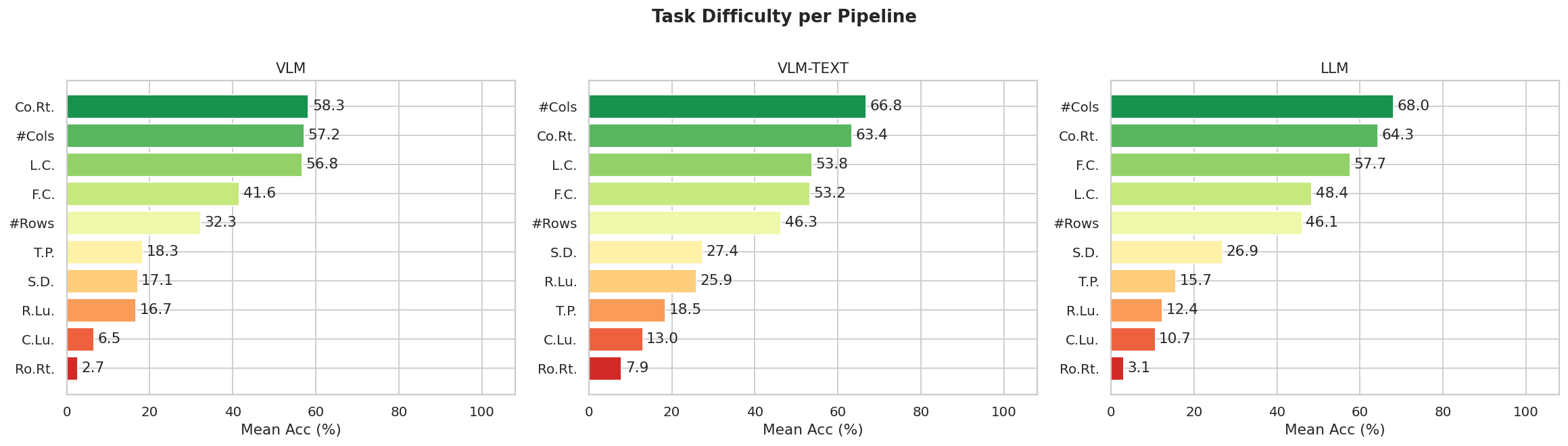}
  \caption{\textbf{Pipeline comparison.} We average exact-match accuracy per SUC subtask for VLM-Image, VLM-Text, and LLM-Text. Higher is better.}
  \label{fig:suc_pipeline_comparison}
\end{figure*}

\paragraph{Pipeline comparison:}
Figure~\ref{fig:suc_pipeline_comparison} compares SUC accuracy across VLM-Image, VLM-Text, and LLM-Text. Structured text generally improves SUC, especially on subtasks that depend on row boundaries and header handling. This is also visible in Tables~\ref{tab:suc_vlm_image}, \ref{tab:suc_vlm_image_appendix}, and~\ref{tab:suc_vlm_text}.
For example, several models improve on row retrieval, reverse lookup, and size detection when the table structure is provided as text.
However, text input does not remove the bottleneck entirely.
Cell lookup and row retrieval remain difficult across pipelines, showing that SUC requires more than access to explicit table markup.

\paragraph{Field accuracy and relaxed accuracy diagnostics:}
Tables~\ref{tab:suc_field_gap} and~\ref{tab:internvl_diagnostic_suc} show that strict EM can hide partial structural recovery. For multi-field tasks such as table partitioning, size detection, cell lookup, and row retrieval, models may partially recover the target structure. Field Accuracy captures this behavior by scoring individual fields, while Relaxed Accuracy captures outputs that contain the correct answer but fail strict string matching. The large gaps between EM and Field Accuracy show that many errors are incomplete or shifted structural predictions rather than completely unrelated answers. This pattern is consistent across both VLM-Image and VLM-Text pipelines, indicating that structural localization remains a common source of failure even when the correct information is partially recovered.

\paragraph{Prompt explicitness and header handling:}
Table~\ref{tab:prompt_sensitivity_changed_vlm} compares the explicit SUC prompt with an implicit prompt on selected VLM-Image subtasks. The explicit prompt states conventions such as excluding headers for first/last-cell tasks and using 0-indexed row/column coordinates for lookup and retrieval tasks, while the implicit prompt removes these details. The largest differences appear on index-dependent subtasks, especially first-cell detection and table partitioning, suggesting that rendered-table models often confuse header rows with table body rows and adopt different row-indexing conventions. The effect is weaker for last-cell detection, which is less affected by header counting. Reverse lookup shows the opposite trend for some models, where the implicit prompt performs better. These effects are particularly pronounced for first-row-sensitive tasks. This confirms that SUC is sensitive not only to visual structure recognition but also to how models interpret indexing and header conventions. 

\begin{figure*}[!h]
  \centering
  \includegraphics[width=0.95\linewidth]{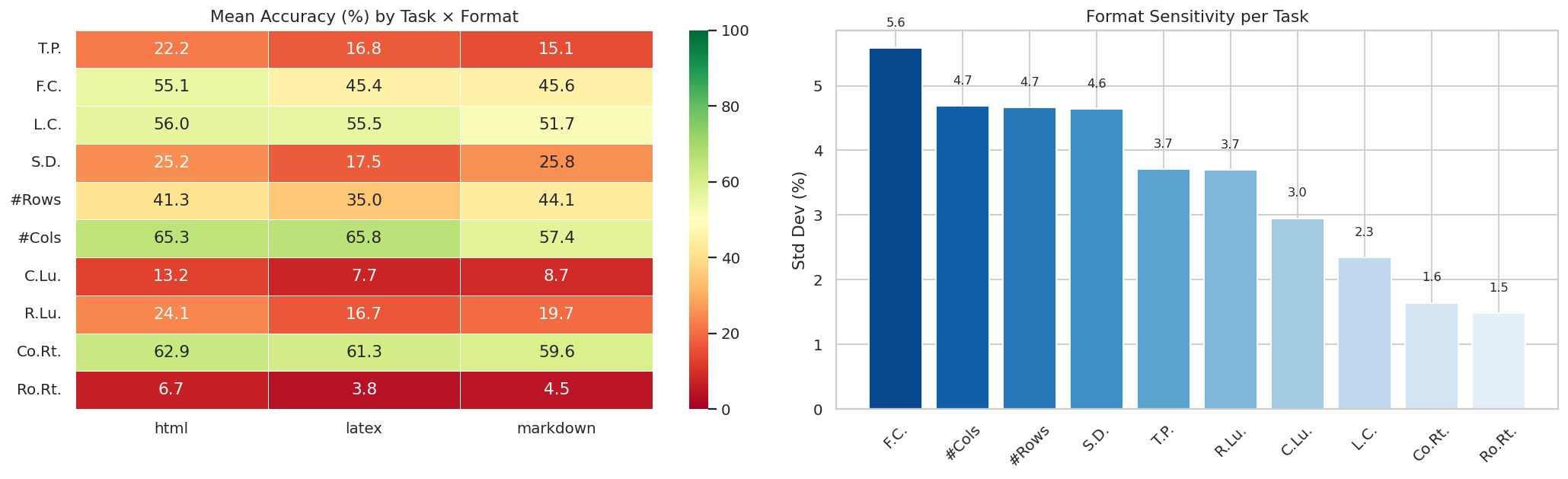}
  \caption{\textbf{Format sensitivity.} We show mean exact-match accuracy by SUC subtask and format (\html{}, LaTeX, \markdown{}) and the variation across formats. Higher is better.}
  \label{fig:suc_format_sensitivity}
\end{figure*}

\paragraph{Format effects:} Figure~\ref{fig:suc_format_sensitivity} and Tables~\ref{tab:suc_vlm_image}, \ref{tab:suc_vlm_image_appendix}, \ref{tab:suc_vlm_text}, and~\ref{tab:suc_llm_text} show that format effects are present but not uniform. For VLM-Image, differences across HTML, LaTeX, and Markdown renders are usually smaller than differences across models and subtasks. Figure~\ref{fig:suc_format_sensitivity} further shows that first-cell detection, column counting, row counting, and size detection are the most format-sensitive subtasks, while row and column retrieval vary little across formats. For text-input pipelines, format effects are more visible. HTML is often the safest structured-text format, especially for VLM-Text and LLM-Text, while LaTeX and Markdown can be less stable for some models. This suggests that SUC depends jointly on input modality, table representation, and row/column conventions.

\begin{table*}[t]
\centering
\small
\setlength{\tabcolsep}{4pt}
\renewcommand{\arraystretch}{0.85}

\definecolor{headerbg}{HTML}{FCE5CD}
\definecolor{openbg}{HTML}{F4CCCC}
\definecolor{closedbg}{HTML}{D9EAD3}

\begin{tabular}{llrrrrr}
\toprule
\rowcolor{headerbg}
\textbf{Model} & \textbf{Format} &
\textbf{EM} & \textbf{Field Acc.} & \textbf{Relaxed Acc.} &
\textbf{$\Delta_{\mathrm{Field}}$} & \textbf{$\Delta_{\mathrm{Relaxed}}$} \\
\midrule

\rowcolor{openbg}
\multicolumn{7}{l}{\textbf{VLM}} \\
\multirow{3}{*}{InternVL3.5-14B}
 & HTML     & 47.8 & 57.5 & 58.3 & +9.7  & +10.5 \\
 & LaTeX    & 46.5 & 57.1 & 57.8 & +10.6 & +11.3 \\
 & Markdown & 48.8 & 59.2 & 59.8 & +10.4 & +11.0 \\
\addlinespace[2pt]
\multirow{3}{*}{InternVL3.5-30B-A3B}
 & HTML     & 38.8 & 52.8 & 53.7 & +14.0 & +14.9 \\
 & LaTeX    & 42.1 & 55.4 & 56.4 & +13.3 & +14.3 \\
 & Markdown & 43.5 & 56.8 & 57.8 & +13.3 & +14.3 \\

\midrule
\rowcolor{closedbg}
\multicolumn{7}{l}{\textbf{VLM-TEXT}} \\
\multirow{3}{*}{InternVL3.5-14B}
 & HTML     & 53.9 & 66.0 & 67.5 & +12.1 & +13.6 \\
 & LaTeX    & 48.9 & 61.2 & 62.4 & +12.3 & +13.5 \\
 & Markdown & 47.9 & 60.8 & 62.5 & +12.9 & +14.6 \\
\addlinespace[2pt]
\multirow{3}{*}{InternVL3.5-30B-A3B}
 & HTML     & 46.9 & 64.2 & 72.0 & +17.3 & +25.1 \\
 & LaTeX    & 42.8 & 57.3 & 60.4 & +14.5 & +17.6 \\
 & Markdown & 42.1 & 57.5 & 65.8 & +15.4 & +23.7 \\

\bottomrule
\end{tabular}

\caption{
\textbf{Diagnostic comparison of InternVL3.5-14B and InternVL3.5-30B-A3B on SUC.}
Values report overall exact-match accuracy (EM), Field Accuracy, and Relaxed Accuracy across formats for the VLM-Image and VLM-Text pipelines.
$\Delta_{\mathrm{Field}} = \mathrm{Field\ Accuracy} - \mathrm{EM}$ and
$\Delta_{\mathrm{Relaxed}} = \mathrm{Relaxed\ Accuracy} - \mathrm{EM}$.
}
\label{tab:internvl_diagnostic_suc}
\end{table*}

\begin{table*}[!h]
\centering
\small
\setlength{\tabcolsep}{3.2pt}
\renewcommand{\arraystretch}{0.82}

\definecolor{headerbg}{HTML}{FCE5CD}
\definecolor{openbg}{HTML}{F4CCCC}
\definecolor{closedbg}{HTML}{D9EAD3}
\definecolor{purplebg}{HTML}{EADCF8}

\begin{tabular}{l c c c c c c c c c c c c}
\toprule
\rowcolor{headerbg}
\textbf{Model} & \textbf{Format} &
\textbf{T.P.} & \textbf{F.C.} & \textbf{L.C.} &
\textbf{S.D.} & \textbf{\# Rows} & \textbf{\# Cols} &
\textbf{C.Lu.} & \textbf{R.Lu.} & \textbf{Co.Rt.} & \textbf{Ro.Rt.} &
\textbf{Overall} \\
\midrule

\rowcolor{openbg}
\multicolumn{13}{l}{\textbf{Open Models}} \\

\multirow{3}{*}{InternVL3.5-14B}
 & HTML     & 8.4 & 95.5 & 78.7 & 8.4 & 50.6 & 98.4 & 50.9 & 44.2 & 95.2 & 8.6 & 53.9 \\
 & LaTeX    & 11.1 & 94.1 & 76.0 & 11.0 & 44.2 & 99.4 & 36.9 & 22.3 & 87.1 & 6.8 & 48.9 \\
 & Markdown & 1.4 & 76.3 & 71.4 & 14.5 & 65.7 & 85.9 & 43.6 & 33.2 & 83.9 & 3.3 & 47.9 \\

\multirow{3}{*}{InternVL3.5-30B-A3B}
 & HTML     & 0.0 & 91.6 & 75.5 & 0.0 & 11.6 & 99.5 & 42.9 & 31.0 & 87.0 & 30.2 & 46.9 \\
 & LaTeX    & 0.0 & 85.1 & 75.8 & 0.0 & 9.1 & 99.4 & 28.8 & 21.0 & 86.8 & 21.9 & 42.8 \\
 & Markdown & 0.0 & 64.4 & 75.4 & 0.0 & 15.7 & 97.1 & 32.4 & 28.5 & 86.2 & 21.5 & 42.1 \\

\multirow{3}{*}{Qwen3-VL-30B-A3B-IT}
 & HTML     & 49.9 & 94.6 & 76.8 & 68.5 & 73.4 & 96.7 & 40.5 & 37.5 & 89.7 & 14.3 & 64.2 \\
 & LaTeX    & 29.4 & 95.5 & 69.6 & 46.9 & 64.5 & 97.5 & 29.7 & 19.7 & 83.0 & 8.6 & 54.5 \\
 & Markdown & 26.4 & 87.8 & 64.9 & 62.3 & 84.6 & 71.7 & 29.6 & 20.8 & 72.2 & 5.9 & 52.6 \\

\multirow{3}{*}{Qwen3-VL-8B-IT}
 & HTML     & 26.9 & 93.5 & 80.0 & 27.8 & 69.0 & 98.3 & 45.3 & 47.9 & 84.9 & 3.2 & 57.7 \\
 & LaTeX    & 18.6 & 92.1 & 76.3 & 15.7 & 36.2 & \textbf{100.0} & 30.2 & 17.5 & 74.1 & 5.6 & 46.6 \\
 & Markdown & 10.7 & 68.5 & 64.5 & 30.2 & 66.9 & 92.8 & 28.1 & 18.8 & 69.0 & 8.3 & 45.8 \\

\multirow{3}{*}{Gemma-3-12B-IT}
 & HTML     & 18.9 & 68.5 & 81.4 & 64.9 & 77.9 & 90.1 & 31.3 & 28.9 & 85.1 & 5.6 & 55.3 \\
 & LaTeX    & 14.5 & 53.1 & 76.5 & 30.5 & 35.6 & 98.7 & 19.1 & 18.4 & 76.9 & 3.7 & 42.7 \\
 & Markdown & 4.3 & 31.6 & 67.4 & 51.4 & 73.3 & 67.2 & 19.4 & 20.7 & 76.3 & 2.5 & 41.4 \\

\multirow{3}{*}{Gemma-3-27B-IT}
 & HTML     & 53.7 & 82.5 & 82.2 & 72.2 & 73.8 & 98.1 & 45.0 & 55.6 & 94.9 & 16.2 & 67.4 \\
 & LaTeX    & 33.2 & 62.6 & 76.2 & 60.4 & 60.1 & 99.5 & 42.0 & 32.3 & 89.0 & 13.2 & 56.9 \\
 & Markdown & 16.4 & 45.3 & 70.6 & 59.5 & 83.3 & 76.5 & 40.5 & 38.5 & 91.6 & 11.6 & 53.4 \\

\multirow{3}{*}{Ministral-3-14B-IT}
 & HTML     & 20.0 & 95.5 & 57.2 & 22.4 & 51.2 & 99.0 & 41.2 & 37.7 & 88.4 & 22.7 & 53.5 \\
 & LaTeX    & 14.1 & 79.0 & 62.3 & 30.2 & 40.5 & 99.4 & 19.2 & 17.0 & 68.5 & 12.2 & 44.3 \\
 & Markdown & 1.7 & 69.0 & 49.4 & 6.5 & 44.7 & 80.0 & 23.8 & 21.3 & 62.0 & 11.1 & 37.0 \\

\multirow{3}{*}{SmolVLM2-2.2B-IT}
 & HTML     & 0.0 & 11.3 & 9.2 & 0.0 & 6.8 & 0.2 & 0.2 & 2.6 & 2.9 & 0.0 & 3.3 \\
 & LaTeX    & 0.0 & 0.5 & 0.2 & 0.0 & 0.0 & 0.0 & 0.0 & 2.7 & 2.6 & 0.0 & 0.6 \\
 & Markdown & 0.0 & 8.3 & 5.9 & 0.0 & 17.9 & 1.8 & 0.0 & 4.5 & 4.2 & 0.0 & 4.3 \\

\multirow{3}{*}{LLaVA-1.6-13B$^*$}
 & HTML     & 0.0 & 21.5 & 5.2 & 1.0 & 12.3 & 31.2 & 2.5 & 0.7 & 16.0 & 0.2 & 9.1 \\
 & LaTeX    & 0.0 & 24.7 & 7.6 & 0.3 & 1.0 & 1.6 & 2.3 & 2.4 & 19.4 & 0.2 & 5.9 \\
 & Markdown & 0.0 & 12.5 & 4.9 & 0.0 & 8.1 & 5.5 & 1.6 & 1.3 & 12.5 & 0.0 & 4.6 \\

\multirow{3}{*}{LLaVA-1.6-7B$^*$}
 & HTML     & 0.0 & 27.5 & 2.7 & 0.0 & 0.0 & 0.0 & 0.0 & 1.0 & 14.4 & 0.0 & 4.6 \\
 & LaTeX    & 0.0 & 38.1 & 2.7 & 0.0 & 10.8 & 8.4 & 1.0 & 0.8 & 13.6 & 0.0 & 7.5 \\
 & Markdown & 0.0 & 16.1 & 2.9 & 0.0 & 1.8 & 0.3 & 0.2 & 0.6 & 3.9 & 0.0 & 2.6 \\

\rowcolor{closedbg}
\multicolumn{13}{l}{\textbf{Table-specialised Models}} \\

\multirow{3}{*}{TableLLaVA-v1.5-7B}
 & HTML     & 0.0 & 19.7 & 2.8 & 0.0 & 7.0 & 0.5 & 0.2 & 1.7 & 7.3 & 0.0 & 3.9 \\
 & LaTeX    & 0.0 & 33.8 & 5.7 & 0.0 & 23.9 & 9.5 & 0.5 & 1.5 & 7.8 & 0.0 & 8.3 \\
 & Markdown & 0.0 & 21.3 & 3.9 & 0.0 & 24.5 & 3.1 & 0.0 & 1.0 & 5.4 & 0.0 & 5.9 \\

\rowcolor{purplebg}
\multicolumn{13}{l}{\textbf{Proprietary Models}} \\

\multirow{3}{*}{GPT-5.2}
 & HTML     & \textbf{94.9} & \textbf{99.4} & \textbf{95.5} & \textbf{98.4} & 97.3 & \textbf{100.0} & \textbf{53.3} & \textbf{86.0} & 94.8 & \textbf{84.4} & \textbf{90.4} \\
 & LaTeX    & 90.5 & 98.4 & 92.5 & 29.3 & 84.4 & \textbf{100.0} & 7.0 & 17.2 & \textbf{97.6} & 3.5 & 62.0 \\
 & Markdown & 88.6 & 94.9 & 93.3 & 87.3 & 96.0 & 99.8 & 14.8 & 34.8 & 96.2 & 25.1 & 73.1 \\

\multirow{3}{*}{Gemini-3-Flash-Preview}
 & HTML     & 92.7 & 97.3 & 90.0 & 86.5 & \textbf{97.8} & \textbf{100.0} & 16.9 & 79.0 & 97.3 & 59.9 & 81.7 \\
 & LaTeX    & 90.6 & 97.1 & 89.3 & 8.9 & 89.8 & \textbf{100.0} & 0.6 & 19.4 & 97.5 & 5.9 & 59.9 \\
 & Markdown & 88.9 & 95.7 & 89.0 & 75.8 & 95.1 & \textbf{100.0} & 15.9 & 67.1 & 97.5 & 49.8 & 77.5 \\

\bottomrule
\end{tabular}

\caption{
SUC results for the VLM-Text pipeline. Exact-match accuracy (\%) across ten structure-oriented subtasks. Models receive table text extracted from HTML, LaTeX, and Markdown sources.}
\label{tab:suc_vlm_text}
\end{table*}

\begin{table*}[t]
\centering
\small
\setlength{\tabcolsep}{5pt}
\renewcommand{\arraystretch}{0.9}

\definecolor{openbg}{HTML}{F4CCCC}
\definecolor{propbg}{HTML}{FCE5CD}
\definecolor{tablebg}{HTML}{D9EAD3}

\begin{tabular}{l c c c c c c c c c c c}
\toprule
\rowcolor{propbg}
\textbf{Models} & \textbf{Formats} &
\textbf{T.P.} & \textbf{F.C.} & \textbf{L.C.} &
\textbf{S.D.} & \textbf{\# Rows} & \textbf{\# Cols} &
\textbf{C.Lu.} & \textbf{R.Lu.} & \textbf{Co.Rt.} & \textbf{Ro.Rt.} \\
\midrule

\rowcolor{openbg}
\multicolumn{12}{l}{\textbf{Open Models}} \\

\multirow{3}{*}{Qwen2.5-7B-Instruct}
 & HTML     & 16.7 & \textbf{95.1} & 57.7 & 20.5 & 52.3 & 80.0 & 25.4 & 14.3 & 75.4 & \textbf{5.7} \\
 & LaTeX    & 11.1 & 44.4 & 54.8 & 3.3 & 34.3 & \textbf{99.0} & 7.3 & 5.2 & 61.4 & 5.4 \\
 & Markdown & 5.9 & 55.6 & 53.7 & 15.1 & 62.5 & 43.7 & 8.1 & 9.2 & 52.6 & 2.7 \\

\multirow{3}{*}{Qwen3-30B-A3B-Instruct}
 & HTML     & \textbf{45.0} & 94.8 & \textbf{68.0} & \textbf{72.2} & \textbf{81.1} & 89.8 & \textbf{44.7} & \textbf{25.0} & \textbf{82.2} & 4.8 \\
 & LaTeX    & 25.6 & 91.3 & 60.9 & 45.5 & 51.2 & 85.4 & 28.3 & 9.4 & 66.9 & 2.2 \\
 & Markdown & 23.4 & 83.5 & 56.0 & 57.6 & 75.5 & 44.4 & 28.9 & 9.2 & 63.4 & 1.4 \\

\rowcolor{tablebg}
\multicolumn{12}{l}{\textbf{Table-specialised Models}} \\

\multirow{3}{*}{TableGPT2-7B}
 & HTML     & \textbf{20.8} & \textbf{94.1} & 68.4 & 31.0 & 49.0 & 78.7 & \textbf{13.5} & \textbf{12.9} & 66.8 & 3.2 \\
 & LaTeX    & 18.1 & 89.7 & \textbf{69.0} & 16.7 & 38.8 & \textbf{98.7} & 6.2 & 7.2 & 57.4 & \textbf{4.1} \\
 & Markdown & 4.9 & 67.2 & 58.0 & \textbf{39.0} & \textbf{66.1} & 38.0 & 6.7 & 9.2 & 49.0 & 1.0 \\

\multirow{3}{*}{TAMA-QWen3}
 & HTML     & 7.2 & 48.5 & 4.9 & 6.8 & 15.1 & 44.2 & 6.4 & 9.4 & \textbf{86.0} & 0.6 \\
 & LaTeX    & 12.1 & 35.1 & 15.9 & 0.0 & 0.0 & 51.0 & 2.1 & 2.1 & 59.3 & 0.2 \\
 & Markdown & 4.6 & 38.0 & 6.0 & 0.0 & 3.2 & 54.2 & 1.7 & 3.5 & 83.6 & 0.3 \\

\bottomrule
\end{tabular}
\caption{
SUC results for the LLM pipeline. Exact-match accuracy (\%) across subtasks is reported.}
\label{tab:suc_llm_text}
\end{table*}

\begin{figure}[!h]
\centering
\includegraphics[
    width=\columnwidth,
    trim=0cm 0cm 0cm 0cm,
    clip
]{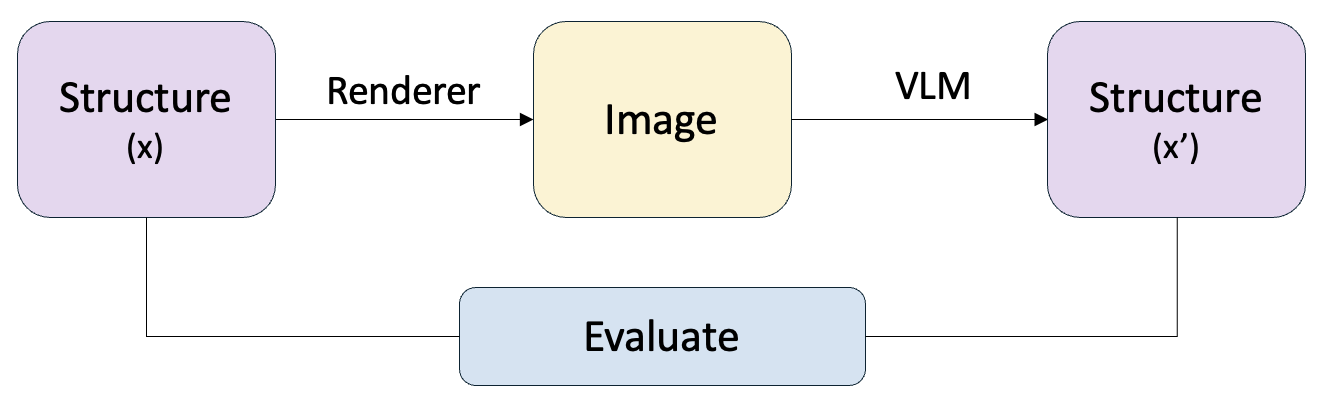}
\caption{Pipeline for SR. A ground-truth table $x$ is rendered into an image, the model predicts a structure $x'$, and evaluation compares $x'$ with $x$.}
\label{fig:generation}
\end{figure}

\paragraph{Evaluation notes:}
We report strict exact-match accuracy as the main SUC metric because the tasks require exact structured outputs, such as row--column coordinates, table size fields, or ordered cell sequences. 

We use the same post-processing across models and formats. Field Accuracy and Relaxed Accuracy are diagnostic metrics for partial structural recovery and formatting errors, but do not replace EM. 

For models with shorter context windows or weaker instruction following, incomplete outputs and formatting errors are counted as incorrect under strict EM to ensure consistent evaluation.

\subsection{Structure Reconstruction: Additional Analyses}
\label{app:sr}

Figure~\ref{fig:generation} illustrates the SR evaluation pipeline, where a table representation is rendered as an image, reconstructed by the model, and compared against the original structure. The analyses below examine reconstruction fidelity, output usability, format-pair difficulty, and cross-format conversion behavior.

\paragraph{Validity-adjusted SR scores:}
Table~\ref{tab:sr_zero_penalized_grits} reports zero-penalized \textsc{GriTS}, where unusable outputs receive a score of zero before averaging. This combines the two SR failure modes: invalid target syntax and inaccurate reconstruction.

The gap between raw and zero-penalized scores is small for strong models on HTML and Markdown targets, showing that most remaining errors are fidelity errors rather than syntax failures. The gap is larger for LaTeX targets, especially for weaker VLMs and TableLLaVA, confirming that LaTeX failures often arise from invalid or non-compilable outputs rather than low structural similarity alone. Combined with the usability results in Table~\ref{tab:sr_output_usability_app}, this indicates that output validity remains a major source of error primarily for LaTeX generation across model families. 

\begin{table*}[t]
\centering
\small
\setlength{\tabcolsep}{0.9pt}
\renewcommand{\arraystretch}{1}

\definecolor{headerbg}{HTML}{FCE5CD}
\definecolor{openbg}{HTML}{F4CCCC}
\definecolor{closedbg}{HTML}{D9EAD3}

\begin{tabular}{l *{18}{c}}
\toprule

\multirow{3}{*}{\textbf{Models}} &
\multicolumn{6}{c}{\textbf{\html{} image}} &
\multicolumn{6}{c}{\textbf{\markdown{} image}} &
\multicolumn{6}{c}{\textbf{\latex{} image}} \\

\cmidrule(lr){2-7}\cmidrule(lr){8-13}\cmidrule(lr){14-19}

& \multicolumn{3}{c}{\textbf{Topology-Zero}} & \multicolumn{3}{c}{\textbf{Content-Zero}}
& \multicolumn{3}{c}{\textbf{Topology-Zero}} & \multicolumn{3}{c}{\textbf{Content-Zero}}
& \multicolumn{3}{c}{\textbf{Topology-Zero}} & \multicolumn{3}{c}{\textbf{Content-Zero}} \\

\cmidrule(lr){2-4}\cmidrule(lr){5-7}
\cmidrule(lr){8-10}\cmidrule(lr){11-13}
\cmidrule(lr){14-16}\cmidrule(lr){17-19}

& \html{} & Md & TeX & \html{} & Md & TeX
& \html{} & Md & TeX & \html{} & Md & TeX
& \html{} & Md & TeX & \html{} & Md & TeX \\
\midrule

\rowcolor{openbg}
\multicolumn{19}{l}{\textbf{Open Models}} \\

SmolVLM2-2.2B-IT & 0.87 & 0.85 & 0.31 & 0.74 & 0.73 & 0.27 & 0.66 & 0.76 & 0.30 & 0.57 & 0.66 & 0.26 & 0.79 & 0.83 & 0.57 & 0.66 & 0.68 & 0.47 \\
Gemma3-12B-IT & 0.94 & 0.95 & 0.75 & 0.79 & 0.79 & 0.63 & 0.95 & 0.96 & 0.75 & 0.80 & 0.81 & 0.63 & 0.94 & 0.94 & 0.74 & 0.78 & 0.77 & 0.62 \\
Gemma3-27B-IT & 0.97 & 0.97 & 0.84 & 0.86 & 0.84 & 0.75 & 0.98 & 0.97 & 0.84 & 0.85 & 0.85 & 0.74 & 0.97 & 0.96 & 0.83 & 0.83 & 0.82 & 0.73 \\
InternVL3.5-14B & 0.99 & 0.99 & 0.89 & 0.95 & 0.94 & 0.86 & 0.99 & 0.98 & 0.87 & 0.93 & 0.93 & 0.83 & 0.96 & 0.97 & 0.89 & 0.93 & 0.91 & 0.87 \\
InternVL3.5-30B-A3B & 0.98 & 0.99 & 0.88 & 0.95 & 0.94 & 0.85 & 0.99 & 0.99 & 0.87 & 0.93 & 0.94 & 0.84 & 0.96 & 0.97 & 0.87 & 0.93 & 0.92 & 0.84 \\
Qwen3-VL-8B-IT & 0.99 & 1.00 & 0.87 & 0.98 & 0.98 & 0.85 & 0.99 & 1.00 & 0.86 & 0.97 & 0.98 & 0.84 & 0.98 & 0.98 & 0.92 & 0.95 & 0.95 & 0.91 \\
Qwen3-VL-30B-A3B-IT & 0.98 & 0.99 & 0.69 & 0.98 & 0.98 & 0.67 & 0.99 & 1.00 & 0.88 & 0.97 & 0.98 & 0.85 & 0.97 & 0.98 & 0.93 & 0.95 & 0.95 & 0.92 \\
Ministral3-14B-Instruct & 0.98 & 0.94 & 0.83 & 0.94 & 0.90 & 0.81 & 0.98 & 0.94 & 0.83 & 0.92 & 0.90 & 0.80 & 0.95 & 0.92 & 0.86 & 0.87 & 0.86 & 0.82 \\
LLaVA1.6-Vicuna-7B & 0.70 & 0.64 & 0.00 & 0.43 & 0.39 & 0.00 & 0.70 & 0.52 & 0.00 & 0.46 & 0.35 & 0.00 & 0.71 & 0.57 & 0.00 & 0.44 & 0.36 & 0.00 \\
LLaVA1.6-Vicuna-13B & 0.66 & 0.80 & 0.00 & 0.46 & 0.51 & 0.00 & 0.62 & 0.74 & 0.00 & 0.45 & 0.50 & 0.00 & 0.65 & 0.74 & 0.01 & 0.46 & 0.49 & 0.01 \\

\addlinespace[4pt]
\rowcolor{headerbg}
\multicolumn{19}{l}{\textbf{Proprietary Models}} \\

GPT-5.2 & 0.98 & 0.98 & 0.78 & 0.97 & 0.94 & 0.75 & 0.98 & 0.99 & 0.91 & 0.96 & 0.97 & 0.88 & 0.98 & 0.97 & 0.91 & 0.94 & 0.94 & 0.88 \\
Gemini-3-Flash-Preview & 0.05 & 0.96 & 0.63 & 0.05 & 0.94 & 0.62 & 0.86 & 0.97 & 0.50 & 0.85 & 0.96 & 0.49 & 0.65 & 0.93 & 0.57 & 0.63 & 0.91 & 0.57 \\

\addlinespace[4pt]
\rowcolor{closedbg}
\multicolumn{19}{l}{\textbf{Table-specialised Models}} \\

TableLLaVA-v1.5-7B & 0.73 & 0.69 & 0.55 & 0.33 & 0.32 & 0.26 & 0.73 & 0.71 & 0.56 & 0.43 & 0.42 & 0.33 & 0.58 & 0.56 & 0.49 & 0.29 & 0.29 & 0.26 \\

\bottomrule
\end{tabular}

\caption{
\textbf{Usability-aware SR scores.}
We report zero-penalized \texttt{GriTS}-Topology and \texttt{GriTS}-Content, where unusable outputs receive a score of zero before averaging, capturing both reconstruction fidelity and output usability.
}
\label{tab:sr_zero_penalized_grits}
\end{table*}

\begin{table*}[!h]
\centering
\small
\setlength{\tabcolsep}{4pt}
\renewcommand{\arraystretch}{1}

\definecolor{tieropen}{HTML}{F4CCCC}     
\definecolor{tierprop}{HTML}{FCE5CD}     
\definecolor{tiertable}{HTML}{D9EAD3}    
\definecolor{deltabg}{HTML}{D9EAF7}      

\begin{tabular}{l *{9}{c}}
\toprule

& \multicolumn{3}{c}{\cellcolor{deltabg}\textbf{\html{} image}} &
\multicolumn{3}{c}{\cellcolor{deltabg}\textbf{\markdown{} image}} &
\multicolumn{3}{c}{\cellcolor{deltabg}\textbf{\latex{} image}} \\

\cmidrule(lr){2-4}\cmidrule(lr){5-7}\cmidrule(lr){8-10}

\rowcolor{tierprop}
\textbf{Models}
& \textbf{\html{}} & \textbf{Md} & \textbf{TeX}
& \textbf{\html{}} & \textbf{Md} & \textbf{TeX}
& \textbf{\html{}} & \textbf{Md} & \textbf{TeX} \\
\midrule

\rowcolor{tieropen}
\multicolumn{10}{l}{\textbf{Open Models}} \\

SmolVLM2-2.2B-IT & 0.97 & 0.90 & 0.35 & 0.75 & 0.81 & 0.37 & 0.92 & 0.92 & 0.60 \\
Gemma3-12B-IT & \textbf{1.00} & 0.99 & 0.81 & \textbf{1.00} & \textbf{1.00} & 0.81 & \textbf{1.00} & 0.99 & 0.80 \\
InternVL3.5-14B & \textbf{1.00} & \textbf{1.00} & \textbf{0.91} & \textbf{1.00} & 0.99 & \textbf{0.89} & \textbf{1.00} & 0.99 & 0.91 \\
InternVL3.5-30B-A3B & \textbf{1.00} & \textbf{1.00} & 0.89 & \textbf{1.00} & \textbf{1.00} & 0.88 & \textbf{1.00} & \textbf{1.00} & 0.89 \\
Qwen3-VL-8B-IT & \textbf{1.00} & \textbf{1.00} & 0.88 & \textbf{1.00} & \textbf{1.00} & 0.87 & \textbf{1.00} & \textbf{1.00} & \textbf{0.95} \\
Ministral3-14B-Instruct & \textbf{1.00} & 0.97 & 0.84 & \textbf{1.00} & 0.97 & 0.84 & \textbf{1.00} & 0.98 & 0.88 \\
LLaVA1.6-Vicuna-7B & 0.98 & 0.76 & 0.00 & 0.99 & 0.68 & 0.00 & 0.99 & 0.73 & 0.00 \\

\bottomrule
\end{tabular}

\caption{
\textbf{Additional SR output usability results.}
Values report the fraction of syntactically usable outputs across source and target formats. Best scores per column are shown in bold.
}
\label{tab:sr_output_usability_app}
\end{table*}

\begin{figure*}[t]
  \centering
  \includegraphics[width=\linewidth]{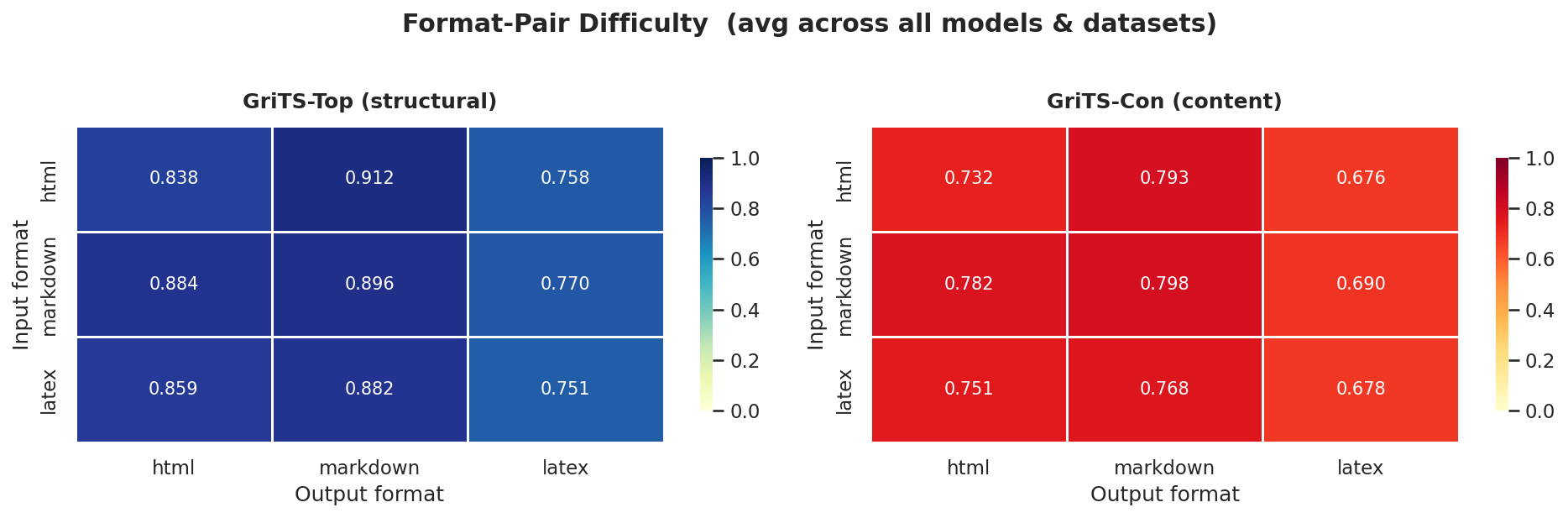}
  \caption{\textbf{Format-pair difficulty (SR).} We average \texttt{GriTS}-Topology and \texttt{GriTS}-Content over all models for each input$\rightarrow$output format pair.}
  \label{fig:sr_format_pair_heatmap}
\end{figure*}

\begin{figure*}[t]
  \centering
  \includegraphics[width=\linewidth]{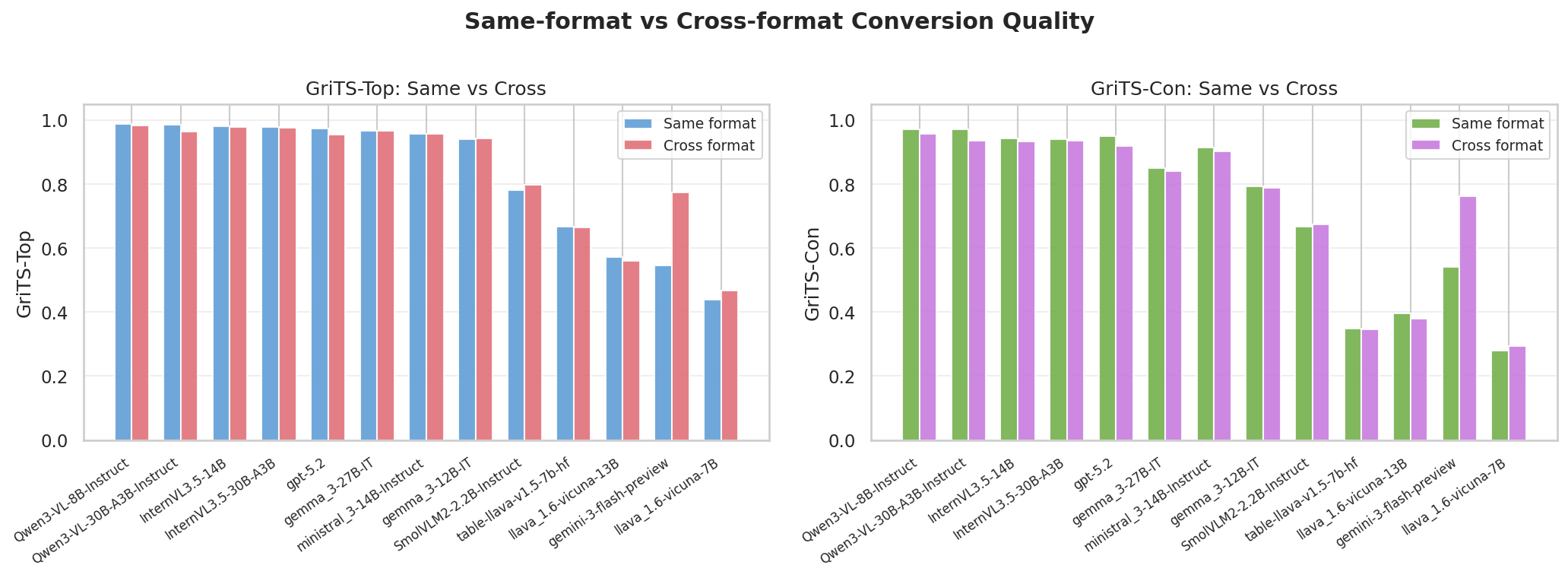}
  \caption{\textbf{Same-format vs cross-format SR.} We compare per-model averages for same-format reconstruction against cross-format conversion.}
  \label{fig:sr_same_vs_cross}
\end{figure*}

\paragraph{Output usability across formats:}
Table~\ref{tab:sr_output_usability_app} reports format-specific usability rates for representative open models. HTML and Markdown outputs are nearly always usable for strong models, with several systems achieving usability close to 100\% across source formats. In contrast, LaTeX usability is consistently lower, even for strong models, and drops to zero for weaker LLaVA variants. These results reinforce the main finding that LaTeX reconstruction is challenging not only because of table structure recovery but also because models must produce syntactically valid target code.

\paragraph{Format-pair difficulty:}
Figure~\ref{fig:sr_format_pair_heatmap} averages \textsc{GriTS}-Topology and \textsc{GriTS}-Content across all models for each input--output format pair. The heatmaps show that output format has a stronger effect than input render format. Markdown targets achieve the highest average scores, while LaTeX targets consistently achieve the lowest. The similarity of rows within each heatmap suggests that models are generally robust to source rendering. Thus, SR difficulty is driven more by the target than the input representation.

\paragraph{Same-format versus cross-format reconstruction:}
Figure~\ref{fig:sr_same_vs_cross} compares same-format reconstruction with cross-format conversion for each model. Strong models such as Qwen3-VL and InternVL3.5 exhibit only small differences between the two settings, indicating that cross-format conversion introduces little additional difficulty once the table structure has been recovered. 

In several cases, cross-format performance is comparable to or slightly better than same-format reconstruction. Larger gaps appear for weaker models, where failures are dominated by unstable parsing, content degradation, or invalid output generation rather than format conversion itself. This suggests that SR errors primarily arise from table understanding and target-format generation, not from translating between representations.

\section{Error Analysis}
\label{sec:erroranalysis}

\begin{table*}
\centering
\small
\setlength{\tabcolsep}{2pt}
\begin{tabular}{p{0.25\linewidth} p{0.55\linewidth} p{0.18\linewidth}}
\toprule
\textbf{Error type} & \textbf{What it looks like} & \textbf{Common in} \\
\midrule
Wrong table cell &
Prediction matches a plausible cell/header, but it comes from the wrong row/column (often satisfies only part of the condition). &
Lookup / Conditional / Comparison \\
\addlinespace
Off-table answer &
Prediction does not match any table cell and does not equal the gold (e.g., generic “cannot determine” replies or free-form values). &
VLM-Image (more) \\
\addlinespace
Multi-item mismatch &
Missing items, extra items, or mixed sets when the question expects a specific set of values. &
Multi-Item Lookup \\
\addlinespace
Multi-item formatting &
Gold items appear in the output, but separators/punctuation break set matching (e.g., commas that belong to an entity name). &
Multi-Item Lookup \\
\addlinespace
Arithmetic near-miss &
Wrong count/sum; off-by-one errors are common when one row is missed or double-counted. &
Aggregation/Arithmetic \\
\addlinespace
Binary label flip (0/1) &
Model outputs a valid 0/1 token but flips the label relative to gold. &
Verification \\
\addlinespace
Answer not isolated &
Gold answer appears in the output but not as the first token (e.g., extra prefix or short descriptor). &
Smaller models; LLM-Text \\
\bottomrule
\end{tabular}
\caption{Common TaskQA failure modes under our exact-match scoring (light normalization; set match for Multi-Item Lookup).}
\label{tab:taskqa_error_types}
\end{table*}

\paragraph{Error analysis (TaskQA):} Table~\ref{tab:taskqa_error_types} summarizes the most common TaskQA failure modes under exact-match evaluation. Most errors arise from incorrect table grounding, incomplete answer retrieval, or answer-format mismatches rather than completely unrelated predictions or random hallucinations. These patterns are consistent across models, formats, and input modalities.

On lookup-style questions, many errors arise from selecting a plausible cell from the correct column but the wrong row, or from missing a filtering condition in Conditional Lookup. Similar failures occur in Comparison/Extremum questions when models compare values within the wrong subset. Multi-Item Lookup introduces two distinct failure modes: missing or extra items, and formatting mismatches that break set matching after normalization. For Aggregation/Arithmetic, errors are typically small counting or summation mistakes caused by skipped or double-counted rows. Verification errors usually appear as 0/1 label flips, often involving negation or multi-row conditions. Exact match also penalizes answers containing extra text before the answer token, a failure mode that is more common for smaller models and explanation-oriented outputs.

\paragraph{Error analysis (SUC):} Table~\ref{tab:suc_error_types} summarizes the main SUC failure modes under strict exact match across all evaluated pipelines. 

We inspected outputs from strong proprietary models (GPT-5.2, Gemini-3-Flash-Preview), strong open models (Qwen3-VL-8B), and a table-specialized baseline (TableLLaVA-v1.5-7B) across VLM-Image, VLM-Text, and LLM-Text settings. We apply identical post-processing and exact-match scoring to all models, making indexing, formatting, and convention mismatches visible.

\begin{table*}[t]
\centering
\small
\setlength{\tabcolsep}{4pt}
\renewcommand{\arraystretch}{1.05}
\begin{tabular}{p{0.23\linewidth} p{0.52\linewidth} p{0.21\linewidth}}
\toprule
\textbf{Error type} & \textbf{What it looks like} & \textbf{Often affects} \\
\midrule
Header offset (\(+1\) row) &
Model treats the header row as part of the indexed grid (header becomes row 0), shifting row indices by 1. &
S.D., \#Rows, C.Lu., Ro.Rt. \\

Repeated-value ambiguity &
The queried value appears multiple times; the model returns a different valid coordinate or lists several coordinates. &
C.Lu., R.Lu. \\

Answer-template mismatch &
Extra text, multi-line answers, or a delimiter different from the required template (e.g., “Row = 3, Col = 2”). &
C.Lu., S.D., \#Rows, Ro.Rt. \\

Row serialization drift &
Row retrieval returns the right row content but with small formatting differences (e.g., leading/trailing `|`, missing cells, or Markdown-style rows). &
Ro.Rt. \\

Verbose structured outputs &
Model outputs a full Markdown table or explanation instead of a single required answer span. &
Mostly Ro.Rt. (also C.Lu./S.D.) \\
\bottomrule
\end{tabular}
\caption{Common SUC error types under strict exact match.}
\label{tab:suc_error_types}
\end{table*}

\paragraph{Header offset explains many indexing/lookup errors, especially in VLM-Image:} Our gold labels define indices over \emph{data cells} (headers excluded).
Under a fixed coordinate convention, some models treat the header as part of the indexable grid, which produces a consistent \(+1\) row shift.
When this happens, size-related probes report one extra row (e.g., gold S.D.\ = \(10|3\) vs.\ prediction \(11|3\)), coordinate probes return row indices that are one larger than gold, and row retrieval often returns the row \emph{one position above} the gold row for the same queried index.

This pattern makes models look strong on boundary and column probes while scoring poorly on C.Lu.\ and Ro.Rt.\ under exact match.
We keep one coordinate convention across pipelines to surface this sensitivity rather than tuning prompts separately per modality or per format.

\paragraph{Repeated values make coordinate probes harder than they appear.}
Many tables contain repeated values (especially short strings and common numbers), so more than one coordinate can look reasonable for C.Lu.\ and R.Lu.
Models then either pick a different occurrence or return multiple coordinates.
Exact match counts both cases as incorrect, even when the output remains consistent with the table content.

\paragraph{Ro.Rt.\ often fails on output formatting, not only row selection.}
Ro.Rt.\ requires emitting the full row as a pipe-separated string.
Models sometimes add leading/trailing bars, change spacing, omit empty cells, or output Markdown-style rows.

TableLLaVA also tends to produce multi-line structured outputs instead of a single row span.
These differences fail exact match even when the chosen row is close.

\paragraph{Verbosity and uniform post-processing.}
Some models (especially TableLLaVA, and occasionally smaller VLMs) generate verbose answers that include the correct information inside extra text or inside a larger structured block.
We intentionally keep one uniform post-processing rule for all models to evaluate end-to-end, machine-readable reliability.
This choice can undercount models that do not follow the requested answer template without using model-specific extraction rules.

\end{document}